\documentclass[10pt,twocolumn]{article}

\usepackage{times}
\usepackage{epsfig}
\usepackage{graphicx}
\usepackage{amsthm,amsmath}
\usepackage{amssymb, verbatim}
\usepackage{fullpage}
\usepackage[ruled,linesnumbered]{algorithm2e}
\usepackage{textcomp} 
\usepackage{multirow}

\begin{document}

\title{Finite Element Based Tracking of Deforming Surfaces\thanks{A preliminary version of this result appeared in 3DIMPVT 2012 [S. Wuhrer, J. Lang, C. Shu, Tracking Complete Deformable Objects with Finite Elements, 3DIMPVT 2012].}}

\author{Stefanie Wuhrer\thanks{Saarland University, Germany, swuhrer@mmci.uni-saarland.de} \and Jochen Lang\thanks{University of Ottawa, Canada} \and Motahareh Tekieh\footnotemark[3] \and Chang Shu\thanks{National Research Council of Canada}}

\date{}

\maketitle

\begin{abstract}
We present an approach to robustly track the geometry of an object that deforms over time from a set of input point clouds captured from a single viewpoint. The deformations we consider are caused by applying forces to known locations on the object's surface. Our method combines the use of prior information on the geometry of the object modeled by a smooth template and the use of a linear finite element method to predict the deformation. This allows the accurate reconstruction of both the observed and the unobserved sides of the object. We present tracking results for noisy low-quality point clouds acquired by either a stereo camera or a depth camera, and simulations with point clouds corrupted by different error terms. We show that our method is also applicable to large non-linear deformations.
\end{abstract}


\section{Introduction}
\label{sec:introduction}

\begin{figure*}[t]
\centering
\includegraphics[width = 0.2\textwidth]{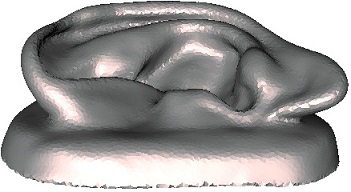}
\includegraphics[width = 0.2\textwidth]{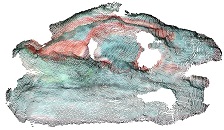}
\includegraphics[width = 0.2\textwidth]{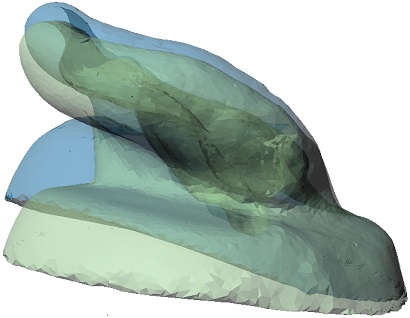}
\includegraphics[width = 0.2\textwidth]{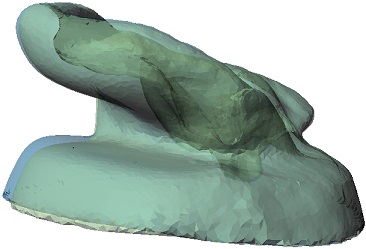}
\caption{\textit{\small Influence of FEM step and comparison to Wuhrer et al. (2012). From left to right: template model, input stereo data (last frame), comparison of the result by Wuhrer et al. (2012) (blue) to our result (green), and comparison of a surface-based deformation (blue) to our result (green). Input models and comparisons shown from different view points.}}
\label{ear_comparison}
\end{figure*}

The accurate acquisition of physical objects has numerous applications, especially in the entertainment industry. While there exist commercial systems for digitizing rigid objects, the acquisition of deforming objects remains a challenge due to the complex changes in geometry over time. A rigid object can be scanned sequentially from multiple viewpoints to accurately capture the complete surface, whereas scanning the entire surface of a deforming object would require a complex and expensive physical setup involving multiple synchronized sensors, which may still be subject to occlusions.

Recently, several techniques were proposed that solve this problem by using a template shape as a geometric and topological prior for the reconstruction and by deforming the template to fit to the observed data~\cite{deAguiar_et_al_08,vlasic_etal_08,li09robust,cagniart2010CVPR}. In some of these methods, the observed data comes from a set of single-view scans. The single viewpoint assumption greatly simplifies the acquisition process. Template-based tracking approaches are shown to lead to visually pleasing results for numerous examples. However, the deformation of the unobserved side of the object is generally only guided by a smoothness function.

We combine a tracking-based approach with fitting a volumetric elastic model to improve the estimation of the unobserved side of the object. We employ a linear finite element method (FEM) to solve for
physical deformations when a given force is applied. Our method proceeds in two steps: First, we use a tracking approach to deform the template model. Second, we use the displacements of the observed vertices of
the template mesh found using the tracking step in a FEM to predict the displacements of the unobserved vertices. Hence, rather than smoothly deforming the unobserved side of the model, we deform the unobserved side through the volumetric mesh of the FEM model. We repeatedly linearize the deformation in the FEM at its current deformation state. Note that our method allows for tracking data acquired using single, multiple, or moving viewpoints.

While deformable models have been introduced to computer vision and computer graphics 30 years ago~\cite{terzopoulos_82}, here we combine modern non-rigid template-based tracking with a volumetric elastic model for completion of the deformation at the unobserved side only. Our major contributions are therefore: 
\begin{itemize}
	\item The combination of a non-rigid template-based tracking approach with a linear finite element model to robustly track the complete geometry of an object whose deformation is captured from a single viewpoint only.
	\item The use of a FEM-based model to deform the unseen side leading to more physically plausible results than by using a smoothness cost in the template-based tracking.
	\item Tracking linear and non-linear deformations by repeatedly linearizing the FEM model at its current deformation state.
\end{itemize}

This paper presents the following three major improvements over the preliminary version of this work~\cite{wuhrer_lang_shu_3dimpvt12}. First, we present a computationally more efficient energy formulation to track the deformable object using a non-rigid iterative closest point framework. Second, we propose an iterative method for the FEM estimation that considers forces at supporting surfaces of the model. Our method does not require the measurement of forces but estimates the forces up to a scale factor from the FEM model and the deformation. Third, we evaluate the performance of our method extensively using numerous synthetic and scanned data sequences and compare our results to our preliminary findings. Special attention is paid to the influence of both synthetic and real scanner noise, as well as to the influence of the FEM step on the result. 

\section{Related Work}
\label{sec:related}

This section reviews work related to tracking surfaces and predicting shape deformations using finite element models. 

\subsection{Tracking}
Computing the correspondence between deformed shapes has received considerable attention in recent years and the surveys of van Kaick et al.~\cite{vanKaick_egstar10} and Tam et al.~\cite{tam_survey_2013} give a comprehensive overview of existing methods. The review in this paper focuses on techniques that do not employ a priori skeletal models or manually placed marker positions, as we aim to minimize assumptions about the structure of the surface. None of the following approaches combine physics-based models with tracking approaches.

\subsubsection*{Template-based approaches}

The following techniques solve the tracking problem using a template as shape prior. De Aguiar et al.~\cite{deAguiar_et_al_08} tracked multi-view stereo data of a human subject that is acquired using a set of cameras. The algorithm uses a template of the human subject that was acquired in a similar posture as the first frame. The approach used for tracking first uses Laplace deformations of a volumetric mesh to find the rough shape deformation and refines the shape using a surface deformation. The deformation makes use of automatically computed features that are found based on color information. Vlasic et al.~\cite{vlasic_etal_08} developed a similar system to track multi-view stereo data of human subjects. Tung and Marsuyama~\cite{tung_matsuyama_10} extended de Aguiar et al.'s approach by using 3D shape rather than color information to find the features.

Li et al.~\cite{li08corres, li09robust} proposed a generic data-driven technique for mesh tracking. A template is used to model the rough geometry of the deformed object, and the algorithm deforms this template to each observed frame. A deformation graph is used to derive a coarse-to-fine strategy that decouples the complexity of the original mesh geometry from the representation of the deformation. 
Cagniart et al.~\cite{cagniart2010CVPR} proposed an alternative where the template is decomposed into a set of patches to which vertices are attached. The template is then deformed to each observed frame using a data term that encourages inter-patch rigidity. Cagniart et al.~\cite{cagniart2010ECCV} extended this technique to allow for outliers by using a probabilistic framework. 

In this work, we combine a template fitting method with a finite element step.

\subsubsection*{Template-free approaches}

The following techniques solve the tracking problem without using a shape prior. However, the methods assume prior information on the deformation of the object. Mitra et al.~\cite{mfoggp_dyn_reg_07} modeled the surface tracking problem as a problem of finding a smooth space-time surface in four-dimensional space. To achieve this, they exploited the temporal coherence in the densely sampled data with respect to both time and space. 
Sharf et al.~\cite{sharf_et_al_08} used a similar concept to find a volumetric space-time solid. Their approach assumes that each cell of the discretized four-dimensional space contains the amount of material that flowed into it.

Wand et al.~\cite{wand_et_al_07} used a probabilistic model based on Bayesian statistics to track a deformable model. The surface is modeled as a graph of oriented particles that move in time. The position and orientation of the particles are controlled by statistical potentials that trade off data fitting and surface smoothness. Tevs et al.~\cite{tevs_etal_12} extended this approach by first tracking a few stable landmarks and by subsequently computing a dense matching.

Furukawa and Ponce~\cite{furukawa_ponce_08} proposed a technique to track data from a multi-camera setup. Instead of using a template of the shape as prior information, their technique computes the polyhedral mesh that captures the first frame and deforms this mesh to the data in subsequent frames. 

Liao et al.~\cite{liao_etal_2009} took images acquired using a single depth camera from different viewpoints while the object deforms and assembled them into a complete deformable model over time. 
Popa et al.~\cite{popa10corres} used a similar approach that is tailored to allow for topological consistency across the motion.

Li et al.~\cite{li12corres} avoid the use of a template model by initializing the tracking procedure with the visual hull of the object. 
Zheng et al.~\cite{zheng_et_al_10} track a deformable model using a skeleton-based approach, where the skeleton is computed using the data. 

\subsection{Predicting Shape Deformations Using FEM}

Several authors suggested learning the parameters of linear finite element models from a set of observations. We use such a method in combination with a tracking method to find an accurate tracking result of the observed and the unobserved side of the model. For a summary of linear finite element methods for elastic deformations, refer to Bro-Nielsen~\cite{bro-nielsen_fem_98}.

Lang and Pai~\cite{lang_pai_01} used a surface displacement and contact force at one point along with range-flow data to estimate elastic constants of homogeneous isotropic materials using numerical optimization.
Becker and Teschner~\cite{becker_teschner_07} presented an approach to estimate the elasticity parameters for isotropic materials using a linear finite element method. The approach takes as input a set of displacement and force measurements and uses them to compute the Young's modulus and Poisson's ratio (see Bro-Nielsen~\cite{bro-nielsen_fem_98}) with the help of an efficient quadratic programming technique. Eskandari et al.~\cite{eskandari_etal_11_viscoelasticFEM} presented a similar approach to estimate both the elasticity and viscosity parameters for viscoelastic materials using a linear finite element method. This approach reduces the problem to solving a linear system of equations and has been shown to run in real-time.

Syllebranque and Boivin~\cite{syllebranque_boivin_08} estimated the parameters of a quasi-static finite element simulation of a deformable solid object from a video sequence. The problems of optimizing the Young's modulus and the Poisson ratio were solved sequentially.
Schnabel et al.~\cite{schnabel_etal_03} used finite element models to validate the non-rigid image registration of magnetic resonance images. Nguyen and Boyce~\cite{nguyen_boyce_10} presented an approach to estimate the anisotropic material properties of the cornea. 

Bickel et al.~\cite{bickel_etal_09} proposed a physics-based approach to predict shape deformations. First, a set of deformations is measured by recording both the force applied to an object and the resulting shape of the object. This information is used to learn a relationship between the applied force and the shape deformation, which allows to predict the shape deformation for new force inputs. The technique assumes the object material to have linear elasticity properties. Bickel et al.~\cite{bickel_etal_10} extended this approach to allow the modeling and fabrication of materials with a desired deformation behavior using stacked layers of homogeneous materials.

Recently, Choi and Szymczak~\cite{choi_szymczak_09} used FEM to predict a consistent set of deformations from a sequence of coarse watertight meshes but their method does not apply to single-view tracking, which is the focus of our method.

Finite element models are used in medical applications to estimate the material parameters of different tissues because the stiffness of a particular tissue can be used to detect anomalies~\cite{hensel_etal_2007_multiorganFEM, zhu_etal_2003_youngsModulusReconstruction, gao_etal_1996_elasticTissuesReview}. For instance, elasticity information can help to segment medical data~\cite{hensel_etal_2007_multiorganFEM} or to detect cancerous tissues~\cite{zhu_etal_2003_youngsModulusReconstruction} or malignant lesions~\cite{gao_etal_1996_elasticTissuesReview}. In this context, Lee et al.~\cite{lee-tmi_2012} proposed a method to estimate material parameters and forces of tissue deformations based on two given frames of observations. This method, which is similar in spirit to our approach, assumes that for both frames, the segmented boundaries of the organ (which are 3D surfaces) are given. The approach proceeds by repeatedly simulating a deformation from the start frame for a set of material parameters and input forces, and by measuring the distance of the simulated deformation to the given target surface. The distance to the target surface is then used to improve the estimated material parameters and input forces using a gradient descent technique. Unlike our method, this approach operates exclusively on a tetrahedral volumetric mesh and therefore has limited resolution. A more serious limitation of the approach is the need for good initial material parameters and force directions. While good initial estimates are known for many organic tissues, we do not have access to good initial estimates in our application. Hence, our approach takes a different strategy to optimize the material parameters that does not require initial estimates.

\section{Overview}
\label{sec:overview}

The input to our method consists of a closed template $T$, a set of contact points on $T$ along with force directions that lead to the deformation of the model, and a set of observed 3D video frames (point clouds) $F_1, F_2, \ldots, F_n$ capturing the deformation. We assume that the template has roughly the shape of the object before the deformation, and is approximately aligned to the first frame.

\begin{figure}[t]
\centering
\includegraphics[width = 7.5cm]{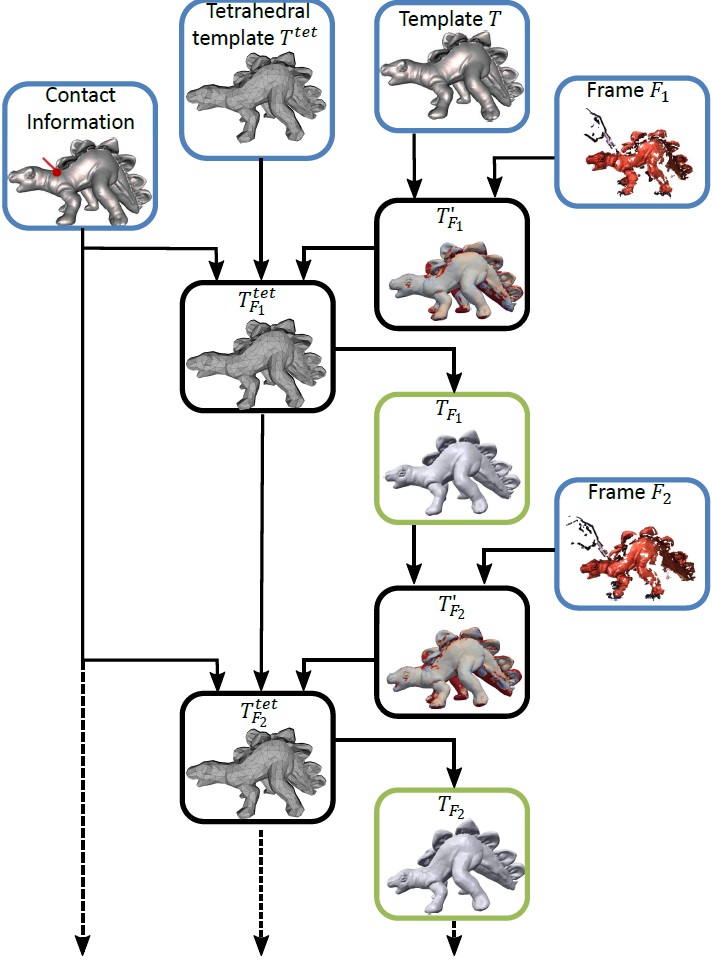}
\caption{\textit{\small Visual overview of the proposed method. Blue boxes show input to the method, black boxes show intermediate results, and green boxes show final results.}}
\label{fig_overview}
\end{figure}

The main idea of our approach is to combine tracking the observed point cloud data using a template deformation approach and predicting the deformation on the unobserved side of the model using a linear FEM. To track the data, we use an energy optimization approach that aims to find deformation parameters that deform $T$ to be close to the observed data and that maintain the geometric features of $T$. Let $T_{F_j}$ denote the template that was deformed to fit to $F_j$. Afterwards, we displace the vertices of $T_{F_j}$ that are not observed in the data with a linear FEM using the given contact point and force direction. To compute the FEM deformation, we use a down sampled version of $T$ that is tetrahedralized, which is denoted by $T^{tet}$ in the following. Let $T^{tet}_{F_j}$ denote the deformed tetrahedral mesh. Finally, we readjust the shape of the unobserved side of $T_{F_j}$ to take this displacement into account. When multiple frames are recorded, we start the tracking and FEM deformation for frame $F_{j+1}$ from $T_{F_j}$ and $T^{tet}_{F_j}$, respectively. Fig.~\ref{fig_overview} gives a visual overview of the proposed method. 

We use the following representation for the deformation of $T$. Let $p_i$ denote the vertices of $T$, $\vec{p}_i$ denote their position vectors, and $\tilde{p}_i$ denote their homogeneous coordinates. Furthermore, let $\vec{n}_{p_i}$ denote the unit outer normal at $p_i$. We deform $T$ by applying a $3\times 4$ transformation matrix $A_i$ to each $\tilde{p}_i$ of $T$. The transformation matrix $A_i$ depends on six parameters: three parameters $\vec{t}_i$ describing a translation, two parameters $\vec{r}_i$ describing a unit rotation axis, and one parameter $\phi_i$ describing a rotation angle. That is, $A_i$ describes a rigid transformation as 
{\small
\begin{equation}
A_i = A_{trans}(\vec{p}_i) A_{rot}(\vec{r}_i, \phi_i) A_{trans}(\vec{t}_i) A_{trans}(-\vec{p}_i), 
\label{deformation_matrix}
\end{equation}
}
where $A_{trans}(-\vec{p}_i)$ is a coordinate transformation that expresses a point in a coordinate frame centered at $p_i$ and $A_{rot}(\vec{r}_i, -\phi_i)$ is a coordinate transformation that expresses a point in a coordinate frame rotated by angle $\phi_i$ around axis $\vec{r}_i$. Expressing the transformation in a coordinate system centered at $p_i$ has the advantage that differences between the transformations of neighboring vertices can be measured directly.

\section{Tracking of Point Cloud Data}
\label{sec:tracking}

This section presents our energy optimization approach to find the transformation parameters $\vec{t}_i, \vec{r}_i$, and $\phi_i$ that lead to a mesh that is close to the point cloud data.

\subsection{Rigid Alignment}

We aim to deform the template $T$ to the observed data. When fitting $T$ to the first frame, we start by deforming $T$ using a global rigid transformation $A$ to fit the observed data as much as possible. We consider all vertices $p_i$ of $T$ and compute the nearest neighbor $N(p_i)$ of the deformed point $A\tilde{p}_i$ in the point cloud data. Let $\vec{n}_{N(p_i)}$ denote the unit outer normal at $N(p_i)$ in the point cloud. 

To rigidly align $T$ to $F$, we find $A$ by minimizing
{\small
\begin{equation}
E_{initial} = \sum_{i} \omega_i \left\langle A\tilde{p}_i - N(p_i), \vec{n}_{N(p_i)}\right\rangle^2
\end{equation}
}
with respect to seven degrees of freedom (one for scaling, three for rotation, three for translation), where $\omega_i$ is a weight term and $\left\langle \cdot,\cdot \right\rangle$ denotes the scalar product. Note that the term $\left\langle A\tilde{p}_i - N(p_i), \vec{n}_{N(p_i)}\right\rangle$ measures the distance of the transformed point $A\tilde{p}_i$ to the supporting plane of its nearest data point $N(p_i)$, which leads to a faster convergence rate of the algorithm than using the distance from $A\tilde{p}_i$ to $N(p_i)$~\cite{rusinkiewicz_levoy_efficient_icp}. Note that we fix the associated point $N\left(p_i\right)$ of $p_i$ for this step to obtain a differentiable energy function. The scaling term accommodates slight errors in the calibration of the 3D scanner used to acquire the template shape and/or the calibration of the camera system used to acquire the frames. 

The weight $\omega_i$ is used to distinguish valid data observations from invalid ones, and should be one for vertices that have a valid data observation and zero for all other vertices. To exclude data points that are inconsistent with $p_i$, $\omega_i$ is set to zero if the distance between $A\tilde{p}_i$ and $N(p_i)$ is above $dr$ or if the angle between $\vec{n}_{p_i}$ and $\vec{n}_{N(p_i)}$ is above $\alpha$, where $r$ is the average edge length of the undeformed template $T$ and $d$ and $\alpha$ are parameters. 
Setting all of the remaining $\omega_i$ to one has the problem that many vertices of $T$ that are close to the acquisition boundary of $F$ may pick the same nearest neighbor, which leads to poor assignments. To remedy this problem, we wish to set $\omega_i$ to zero if $N(p_i)$ is located on the acquisition boundary of $F$. It is not straight forward to define a ``boundary'' on a noisy point cloud. We use a heuristic that considers points $q_j$ of $F$ to be part of the boundary if many vertices of $T$ choose $q_j$ as nearest neighbor. To find the boundary points in a way that is independent to global resolution changes of both $T$ and $F$, we count for each point of $F$ the number of vertices of $T$ that chose it as nearest neighbor and average this count over all points of $F$ that were chosen by at least one point of $T$. A point of $F$ is then considered a boundary point if its count exceeds twice the average count. 
In all remaining cases, $\omega_i$ is set to one.

\subsection{Non-Rigid Alignment}

To fit $T$ to any frame, we deform $T$ in a non-rigid fashion by changing $\vec{t}_i, \vec{r}_i$, and $\phi_i$ to minimize the energy
{\small
\begin{equation}
\begin{array}{lll}
E_{track} & = & w_{data} E_{data} + w_{sm} E_{sm} \mbox{~with}\\
E_{data} & = & \sum_{i} \omega_i \left\langle A_i\tilde{p}_i - N\left(p_i\right), \vec{n}_{N(p_i)}\right\rangle^2 \\
E_{sm} & = & \sum_{i} \frac{1}{\left|R_{sm}(p_i)\right|} \sum_{j \in R_{sm}(p_i)} (\left\|\vec{t}_i-\vec{t}_j\right\|^2 \\
 & + &\min \left((\phi_i-\phi_j)^2, (2\pi - \left| \phi_i- \phi_j \right| )^2 \right) )\\
\end{array}
\end{equation}
}
where $w_{data}$ and $w_{sm}$ are weights for the individual energy terms, $R_{sm}(p_i)$ is the set of indices corresponding to points $A_j\tilde{p}_j$ that have geodesic distance at most $s_{sm}r$ of $A_i\tilde{p}_i$, and $\left|.\right|$ denotes the cardinality of a set. As before, $r$ is the average edge length of $T$ and $s_{sm}$ is a parameter. As above, $\omega_i$ is set to zero if the angle between $\vec{n}_{p_i}$ and $\vec{n}_{N(p_i)}$ is above $\alpha$, if the distance between $A\tilde{p}_i$ and $N(p_i)$ is above $dr$, or if $N(p_i)$ is a boundary point, and to one otherwise. The transformation matrices $A_i$ are computed according to Equation~\ref{deformation_matrix}.

The data term $E_{data}$ drives the template mesh to the observed data. However, using only this term results in an ill-posed problem. We therefore add the smoothness term $E_{sm}$ to act as a regularization term that encourages smooth transformations. Unlike previously used regularization terms~\cite{allen_curless_popovic_03_parametrization_body_shape,li09robust} that measure the difference between the transformations $A_i$, $E_{sm}$ measures the differences in translations and rotation angles in a local coordinate system centered at $p_i$. To obtain results that are invariant under rigid transformations, it is important to initialize the transformation parameters in a way that is invariant to rigid transformations of the scene. Initializing $\vec{t}_i$ and $\phi_i$ to zero yields the initial identity transformation $A_i$ regardless of how $\vec{r}_i$ is initialized. A natural rotation-invariant choice to initialize $\vec{r}_i$ is $\vec{n}_{p_i}$. If $E_{sm}$ were to depend on the difference in rotation axes with this initialization, $E_{sm}$ would not be zero in the rest state. Hence, the difference in $\vec{r}_i$ does not increase $E_{sm}$. We found that in practice, $E_{sm}$ helps to avoid self-intersections during the deformation, which is important as self-intersections might cause problems in the FEM step.

To minimize $E_{track}$, we start by encouraging smooth transformations by setting $w_{data} = 1$ and $w_{sm} = 100$. Similar to Li et al.~\cite{li09robust}, whenever the energy does not change by much, we relax the smoothness weight as $w_{sm} = w_{sm}/2$ to give more weight to the data term. We stop when the relative change in energy \\
$\left(E_{track}^{(i-1)}-E_{track}^{(i)}\right)/E_{track}^{(i-1)}$, where $i$ is the iteration number, is less than $0.0001$ or when $w_{sm}$ is smaller than $20$. To obtain a differentiable energy function, we do not update the associated point $N\left(p_i\right)$ of $p_i$ for a fixed set of weights. That is, $N\left(p_i\right)$ is updated every time the weight $w_{sm}$ is changed.

\subsection{Multi-Resolution Optimization}

Recall that the distance to the nearest neighbor used in $E_{data}$ is limited by the template resolution. To allow for larger deformations, we use a multi-resolution approach as follows. We compute a multi-resolution hierarchy of $T$ by collapsing edges of the mesh according to Garland and Heckbert's geometry criterion~\cite{garland_heckbert_97}. We do not collapse edges if this would lead to a self-intersecting mesh. We perform the test whether an edge collapse leads to self-intersections greedily by performing the collapse and testing if self-intersections occur. In each resolution step, we halve the number of vertices. We stop the collapse when the base mesh contains about 1000 vertices or when no more valid edges can be found for the edge collapse operation. 

Once $E_{track}$ is minimized for resolution level $l$, we consider the mesh of the next higher resolution level $l+1$. For the vertices of level $l+1$ that are present in level $l$, we initialize the transformation parameters to the result of the previous resolution. For the remaining vertices, $\vec{r}_i$ is initialized to $\vec{n}_{p_i}$, and $\vec{t}_i$ and $\phi_i$ are found by minimizing $E_{sm}$ with respect to the indices $i$ that are not present in resolution level $l$. 

This multi-resolution framework works well when the geometric complexity is approximately linked to the amount of deformation. However, in cases where most of the deformation occurs in feature-less regions of the surface, some deformation detail may be lost by this multi-resolution framework. A possible remedy is to use a deformation graph to compute a multi-resolution framework~\cite{li09robust}.

\subsection{Post-Processing}

While the solution yields a globally smooth deformation field by design, it is not guaranteed to give a deformation field that is locally smooth at every vertex $p_i$. Instead, it may happen that a single vertex is transformed by a significantly different deformation than its neighbors, thereby generating a new feature in the geometry. 

If this happens, we can optionally post-process the result as follows. For every vertex $p_i$ of $T$, we consider the minimum of $\left|A_ip_i-A_jp_j\right| / \left|p_i-p_j\right|$ over all $p_j$ in the one-ring neighborhood of $p_i$. If this minimum is larger than two, which means that the distance of $p_i$ to each of its neighbors has at least doubled during the deformation, we set the transformation parameters of $p_i$ to the average of the transformation parameters of the one-ring neighbors of $p_i$. The average rotation axis is computed using spherical linear interpolation. 

In our experiments, we observed that the post-processing step was usually only required because of the FEM step in the previous frame. That is, the FEM step caused some vertices $p_i$ to move relatively far from their neighbors when updating the unobserved side of the previous frame. This in turn led to a lack of smoothing as $R_{sm}(p_i)$ contained few points. A possible remedy to this problem is to use a more complex multi-resolution framework, as discussed above. We chose not to implement this solution, because in practice, we found that this post-processing was usually not crucial in most examples. In our experiments, less than $1\%$ of the frames were influenced by this post-processing. 

\subsection{Comparison to Prior Work}

Our preliminary work used a data energy based on a point-to-point distance, a smoothness energy that varied depending on the differences in rotation axes, and an energy designed to discourage self-intersections by repelling close-by vertices that are not neighbors~\cite{wuhrer_lang_shu_3dimpvt12}. If we let $M$ denote the complexity to find $N(p_i)$, and assume $R_{sm}(p_i)$ to have constant complexity (which holds for regularly sampled templates), a single evaluation of the energy used by Wuhrer et al. has a complexity of $O(mM + m^2)$ for a template mesh with $m$ vertices as their energy requires the computation of all distances between pairs of vertices on the mesh. In contrast, evaluating $E_{track}$ has a complexity of $O(mM)$. Furthermore, it is known that using a point-to-plane distance instead of a point-to-point distance leads to faster convergence rates~\cite{rusinkiewicz_levoy_efficient_icp}. Hence, we reduced the computational complexity of our method. 

Furthermore, as will be shown in Section~\ref{sec:results}, using $E_{track}$ results in less self-intersections and higher data accuracy than using the energy by Wuhrer et al.

\section{Displacing Unobserved Vertices Using FEM}
\label{sec:unobserved}

Consider the situation after $T$ was deformed to frame $F_j$ using the approach outlined in the previous section, and denote the deformation of $T$ by $T_{F_j}$. We call the vertices $p_i$ in $T_{F_j}$ that were deformed using valid data observations \textit{observed vertices}, and we call the remaining vertices \textit{unobserved vertices}. Unobserved vertices were deformed using smoothness assumptions on the deformation field only. This section describes how to displace the unobserved vertices using a linear FEM. 

\subsection{Linear FEM}

We aim to reposition the unobserved vertices of $T_{F_j}$ using a finite element model. We use $T_{F_{j-1}}$ with $T_{F_0}=T$ as start position for the FEM step. A tetrahedral mesh is used to compute the FEM. The initial tetrahedral mesh of $T$ is obtained by tetrahedralizing a simplified version of $T$. This simplification is necessary to make the algorithm more time and space efficient. The tetrahedral mesh contains vertices on the surface of the model (these vertices are a subset of the vertices of $T_{F_{j-1}}$) and vertices that are internal to the model. In the following, let $T_{F_{j-1}}^{tet}$ denote this tetrahedral mesh. 

The FEM linearly relates the displacements of the vertices and the forces applied to the tetrahedral mesh using a stiffness matrix $K(E, \nu)$ that depends on the geometry of the tetrahedral mesh and on two elasticity parameters, the Young's modulus $E$ and the Poisson ratio $\nu$. Let $\vec{f}$ denote the vector of forces applied to the vertices of the tetrahedral mesh and let $\vec{u}$ denote the vector of displacements of the vertices of the tetrahedral mesh. Both $\vec{f}$ and $\vec{u}$ have dimension $3m$, where $m$ is the number of vertices of the tetrahedral mesh. Furthermore, let $\vec{f}_i$ and $\vec{u}_i$ denote the force and displacement vectors of vertex $i$. Then,
{\small
\begin{equation}
K(E, \nu)\vec{u} = \vec{f}.
\label{fem_equation}
\end{equation}
}

This equation can be used in three ways. 
\begin{itemize}
\item[(A1)] Given all displacements and forces, $E$ and $\nu$ can be estimated by solving a linear system of equations, as shown by Becker and Teschner~\cite{becker_teschner_07}. In principle, if the displacements and forces at all vertices of a single tetrahedron are known, the approach by Becker and Teschner can estimate $E$ and $\nu$. However, due to numerical instabilities when using a single tetrahedron, in practice, redundant observations are commonly used to estimate the material parameters. 
\item[(A2)] Given $E$, $\nu$, and $\vec{u}$, we can compute $\vec{f}$ by a matrix multiplication.
\item[(A3)] Given $E$ and $\nu$ along with at least three fixed displacements, Equation~\ref{fem_equation} can be modified such that $K(E, \nu)$ is invertible. If for each vertex with non-fixed displacement either the force or the displacement are provided, we can compute the missing displacements and force vectors by rearranging the linear system of equations~\cite{bro-nielsen_fem_98}.
\end{itemize}

We rely on forces in addition to displacements in the estimation of unobserved vertices because overspecified boundary conditions are required to estimate material parameters.

Note that this simple linear FEM is only suitable to model small deformations, as large rotations may cause artifacts. However, this problem does not occur in our case as we linearize the deformation locally at each frame by modeling deformations between consecutive frames, which ensures that only small deformations are considered. Because of using the deformed tetrahedral template from the previous tracking frame as the rest state, the material parameters estimated by our method are not expected to be physically meaningful. 

\subsection{Estimating the Positions of Unobserved Vertices}

It is easy to see that we do not have enough constraints to use Equation~\ref{fem_equation} directly to solve for all missing information. From the tracking step, we computed displacements $\vec{u}_i$ for all surface vertices, but not for the internal vertices. These displacements are reliable for observed vertices only, and we aim to use the FEM model to find reliable displacements for the unobserved ones. 

Furthermore, we are given the direction of the force at the contact point. Note that we can normalize the length of the force direction at the contact point, since changing the length of $\vec{f}$ only scales $E$ (see e.g. Becker and Teschner~\cite[Equation 17]{becker_teschner_07}). The forces $\vec{f}_i$ at internal vertices can be assumed to be zero as no external forces can act on the interior of the model. Note that other contact surfaces of the model, such as the table the model rests on, are not modeled explicitly in our framework. Instead, we rely on the observed surface points to model these additional constraints. 

This leaves us with the following unknown or unreliable quantities: $E$, $\nu$, the displacements at internal and unobserved vertices, and the forces at surface vertices that are not contact points.

Prior work assumed all forces to be zero, solved for $E$ and $\nu$ by only considering the points with known displacements and forces, and used the estimated $E$ and $\nu$ to solve for the displacements of unobserved vertices~\cite{wuhrer_lang_shu_3dimpvt12}. This approach does not model all physical constraints as forces from the contact with supporting surface are being set to zero. Hence we did not adopt this approach. 

We propose an iterative method to find reliable displacements at unobserved vertices and demonstrate in Section~\ref{sec:results} that this change leads to an improvement of the tracking results. 

The method starts by using the displacements $\vec{u}_i$ at surface vertices computed using the tracking step as an initial estimate. That is, $\vec{u}_i$ is computed as the difference between the vertex coordinate of $p_i$ on $T_{F_j}$ and its corresponding point on $T_{F_{j-1}}^{tet}$. This estimate is diffused to internal vertices using a thin-plate spline (TPS) deformation. 

The following description of TPS closely follows the description by Dryden and Mardia~\cite[Chapter 10.3]{dryden_mardia_shape_analysis}. Let $P=\left[\vec{p}_1, \ldots, \vec{p}_m\right]$ denote the $3\times m$ matrix of the coordinate vectors of the vertices of $T_{F_{j-1}}$, and let $U=\left[\vec{u}_1,\ldots,\vec{u}_m\right]$ denote the $3\times m$ matrix of displacement vectors sorted in the same order as in $P$. The TPS deformation is $\Phi(\vec{p}) = \vec{c} + X \vec{p} + Y^T s(\vec{p})$, where $\vec{c}$ is a $3$-dimensional vector, $X$ is a $3 \times 3$ matrix, $Y$ is a $m\times 3$ matrix, and $s(\vec{p})=\left[\phi(\vec{p}-\vec{p}_1) \ldots \phi(\vec{p}-\vec{p}_m)\right]^T$ is a $m$-dimensional vector with 
{\small
\begin{equation}
\phi(\vec{p})=\begin{cases}\left\|\vec{p}\right\|^2 \log{\left\|\vec{p}\right\|}, \left\|\vec{p}\right\| > 0,\\
0, otherwise \end{cases}.
\end{equation} 
}
We find $\Phi(\vec{p})$ by solving the linear system of equations
{\small
\begin{equation}
\begin{bmatrix}S & \textbf{1} & P^T \\	\textbf{1}^T & 0 & 0 \\ P & 0 & 0 \\ \end{bmatrix} \begin{bmatrix}Y \\	c^T \\ X^T \\ \end{bmatrix} = \begin{bmatrix} U^T \\ 0 \\ 0 \\ \end{bmatrix},
\end{equation} 
}
where $\textbf{1}$ is a $m\times 1$ vector containing $1$ at each position, and $S$ is a matrix containing the vectors $s(\vec{p}_i)$ as its rows. 

We then evaluate $\Phi(\vec{p})$ at the internal nodes of $T_{F_{j-1}}^{tet}$ to obtain an initial estimate for the displacements. 

Let $\vec{u}_{init}$ denote the vector of length $3m$ containing all initially estimated displacements. These displacements are used to iteratively update the material parameters using (A1) and the estimated forces using (A2). Finally, the estimated $E$, $\nu$ and $\vec{f}$ along with fixed displacements at observed vertices are used to estimate $\vec{u}_i$ at unobserved vertices using (A3). Algorithm~\ref{alg_estimate_fem} summarizes this approach.

\begin{algorithm}[htb]
	\SetAlgoLined
	\KwData{$T_{F_{j-1}}^{tet}$, $F_j$}
	\KwResult{$\vec{u}$}
	Compute $\vec{u}_{init}$ based on $F_j$ and a thin-plate spline deformation\;
	Initialize a set $\mathcal{F}$ of indices corresponding to known forces\;
	\For {$\ell$ iterations}
	{
		Use the known $\vec{f}_i$ to solve for $E, \nu$ by minimizing $\sum_{i\in \mathcal{F}} \left\|(K(E, \nu)\vec{u}_{init})_i-\vec{f}_i\right\|^2$\;
		Solve for the unknown $\vec{f}_i$ using $K\vec{u}_{init} = \vec{f}$\;
		\If{first iteration}
		{
			Update $\mathcal{F}$ to contain all force indices\; 
		}
	}
	Use the estimated $E, \nu, \vec{f}$ to find $\vec{u}_i$ at unobserved vertices\;
	\caption{Using FEM to displace unobserved vertices}
	\label{alg_estimate_fem}
\end{algorithm}

After the FEM step, the tetrahedral mesh of $T_{F_j}$ is obtained by simply updating the vertex positions of $T_{F_{j-1}}$ using $\vec{u}_i$. 

In our experiments, we set $\ell = 3$ (since we found this to be sufficient in our experiments).

\subsection{Updating the Transformation Parameters}

Once the FEM step is completed for frame $F_j$, it remains to adjust the transformation parameters $\vec{t}_i, \vec{r}_i$, and $\phi_i$ to capture the new deformation. We achieve this by minimizing 
{\small
\begin{equation}
\begin{array}{lll}
E_{def} & = & w_{data} E_{target} + w_{sm} E_{sm}\mbox{~with} \\
E_{target} & = & \sum_i \left\|A_i \tilde{p}_i - TP(p_i) \right\|^2, \\
\end{array}
\end{equation}
}
where $TP(p_i)$ is the position of the point corresponding to vertex $p_i$ on the deformed tetrahedral mesh $T_{F_{j-1}}^{tet}$. Note that we only optimize the energy with respect to parameters that influence unobserved vertices of $T_{F_j}$. In our experiments, we set $w_{data}=1$ and $w_{sm}=20$.

\section{Implementation Details}
\label{sec:implementation}

The implementation of the algorithm is in C++ and uses a quasi-Newton method~\cite{liu_nocedal_lbfgsb} for all of the optimization steps. For each optimization step, at most 1000 iterations are used. The tetrahedralization is computed using tetgen (http://tetgen.berlios.de). When tetrahedralizing the model, we find a high quality tetrahedralization by restricting the radius-edge ratio of each tetrahedron to be at most two.

This section discusses implementation details, and in particular the parameter settings used in the experiments. The parameters $w_{data}$ and $w_{sm}$ used during tracking give the relative weights between the different energy terms, the parameters $d$ and $\alpha$ control which data points influence the data term, and the parameter $s_{sm}$ influences the neighborhoods considered for the smoothing term. Finally, the parameter $\ell$ controls how many iterations are performed during the FEM estimation.

\begin{figure*}[tb]
\centering
\begin{tabular}{cccc}
$s_{sm}=5.0$ & $s_{sm}=3.0$ & $s_{sm}=1.5$ & \multirow{2}{*}{\includegraphics[height = 2.3cm]{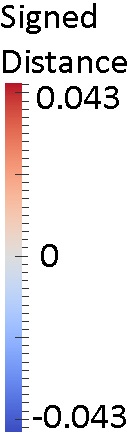}}\\
\\
\includegraphics[height=2.0cm]{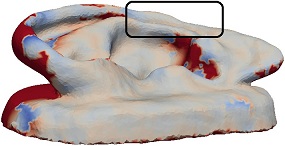} & 
\includegraphics[height=2.0cm]{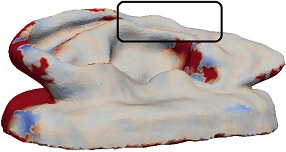} & 
\includegraphics[height=2.0cm]{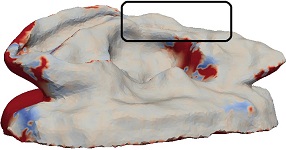} &
\\
\end{tabular}
\caption{\textit{\small Influence of parameter $s_{sm}$. Tracking results are color-coded according to the signed distance from the data.}}
\label{ear_parameter_s_sm}
\end{figure*}

To make the relative influence of the weights $w_{data}$ and $w_{sm}$ invariant with respect to scaling, we pre-scale all of the input models, such that the length of the diagonal of the bounding box of the template model is one. This allows to set most of the parameters to one constant value for all experiments. The weight schedule used for $w_{data}$ and $w_{sm}$ as well as the choice of $\ell$ has been discussed in Sections~\ref{sec:tracking} and~\ref{sec:unobserved}. Furthermore, we set $w_{data}=1, d=5, \alpha=60^{\circ}$. 

The parameter $s_{sm}$ is the only parameter that is varied. This parameter gives the smoothing radius with respect to the resolution of the template mesh. It needs to be adjusted depending on the ratio between the mesh resolution and the mesh size. If the mesh resolution (measured as average edge length) is high compared to the size of the model, then $s_{sm}$ can be set relatively low. If the mesh resolution is low compared to the size of the model, then $s_{sm}$ needs to be set to a higher value. 

Fig.~\ref{ear_parameter_s_sm} shows the influence of $s_{sm}$ on the result of tracking scan data of an ear model. The larger the parameter $s_{sm}$, the more localized shape deformations are penalized by the tracking energy. This has the effect that for small $s_{sm}$, the template can accurately follow the data at the cost of being influenced by data noise and for large $s_{sm}$, the template is not significantly affected by noise but cannot follow localized shape deformations. In our experiments, we set $s_{sm} = 1.5$ for synthetic data, $s_{sm} = 3$ for the ear model, and $s_{sm} = 5$ for the dinosaur model.

\section{Evaluation}
\label{sec:results}

This section discusses the datasets used in the experiments and shows a synthetic evaluation of the method as well as experiments based on real data. Furthermore, we compare the proposed method to its predecessor~\cite{wuhrer_lang_shu_3dimpvt12}, denoted by \textit{Wuhrer et al. (2012)} in the following. More detailed visualizations of some experiments are available in the supplementary material. For all the experiments, the input models are pre-scaled, such that the length of the bounding box diagonal of the template model is one. This information on the scale of the models serves as reference for the numerical evaluations below.

\subsection{Input Data}

\begin{figure}[tp]
\centering
\begin{tabular}{|c|c|c|}
\hline
\includegraphics[height=1.9cm]{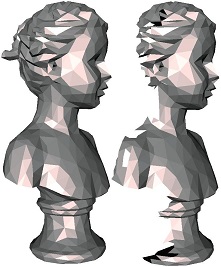} &
\includegraphics[height=1.9cm]{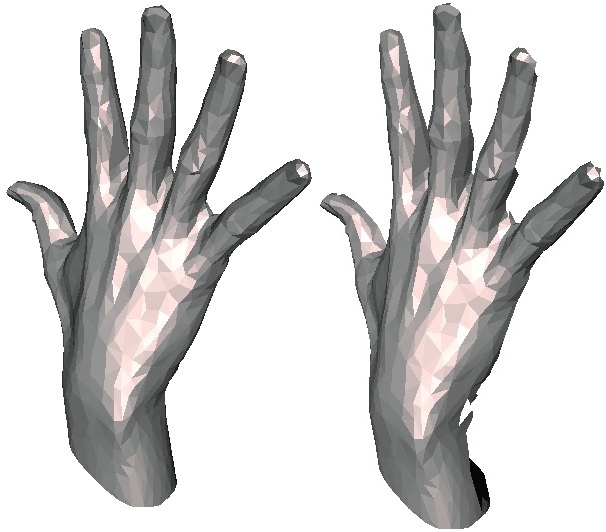} &
\includegraphics[height=1.9cm]{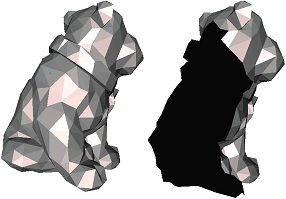}\\
\hline
\end{tabular}
\caption{\textit{\small Bust, hand, bulldog models: model and front side.}}
\label{buste_hand_dog_template}
\end{figure}

\begin{figure}[tp]
\centering
\begin{tabular}{c c c}
\includegraphics[scale=0.3]{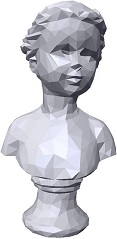} &
\includegraphics[scale=0.3]{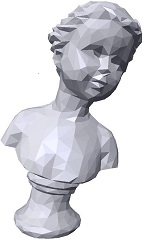} &
\includegraphics[scale=0.3]{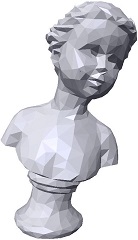}\\
\includegraphics[scale=0.12]{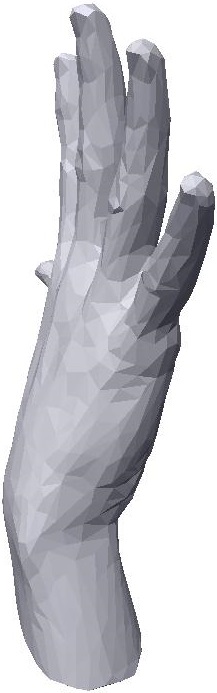} &
\includegraphics[scale=0.12]{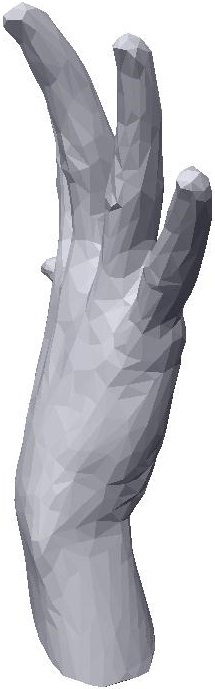} &
\includegraphics[scale=0.12]{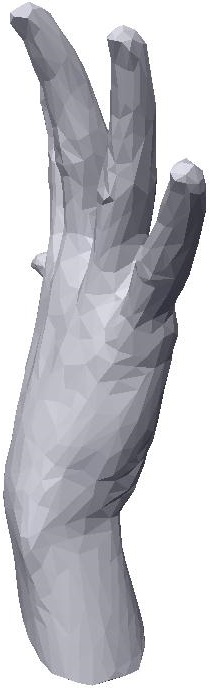}\\
\includegraphics[scale=0.3]{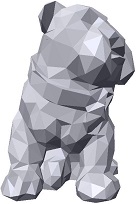} &
\includegraphics[scale=0.3]{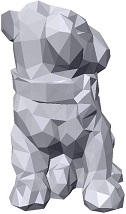} &
\includegraphics[scale=0.3]{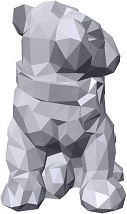}\\
\end{tabular}
\caption{\textit{\small Left to right: start position, end position of linear FEM deformation, end position of StVK deformation.}}
\label{buste_hand_dog_deformations}
\end{figure}

\textbf{Synthetic Data.}  The synthetic datasets (shown in Fig.~\ref{buste_hand_dog_template}) are created using the bust, hand, and bulldog models from the AIM@Shape repository~\cite{aimAtShape}. We create synthetic deformations of the models by applying different finite element deformations to the models with GetFEM~\cite{getfem}. First, the shapes are deformed using a linear FEM, and second, the shapes are deformed using the incompressible non-linear Saint Venant-Kirchhoff model (StVK). The back sides of the deformed models are removed and the remaining front sides (shown in Fig.~\ref{buste_hand_dog_template}) are used as input to the algorithm. In our simulations, the head of the bust is pushed to the left, the middle finger of the hand is pushed to the left, and the head of the bulldog is pushed to the side. For all deformations, the Lagrange multiplier was set to $1e4$. Refer to Table~\ref{info_synthetic} for more information on the models and the parameters used to generate these deformations, and to Fig.~\ref{buste_hand_dog_deformations} for the start and end poses of the deformations. 

\begin{table}[bt]
\centering
{\small
\begin{tabular}{|l|c|c|c|c|}
\hline
Model & Number & Number & $\nu$ & $E$  \\
 & Vertices & Frames & ($\,\mathrm{{N}/{mm^2}}$) & ($\,\mathrm{{N}/{mm^2}}$)\\
\hline
Bust & 7470 & 10 & $0.35$ & $5.999e4$ \\
\hline
Hand & 1257 & 13 & $0.35$ & $5.999e4$ \\
\hline
Bulldog & 847 & 25 & $0.49$ & $2.980e3$ \\
\hline
\end{tabular}
}
\caption{\textit{\small Information about the synthetic models.}}
\label{info_synthetic}
\end{table}

\textbf{Stereo Data.}  We acquire the stereo observations of each frame using a commercial machine vision stereo camera (Point Grey Bumblebee 2) with the window-based matching software of the manufacturer. Typically, we use a matching window of size $15 \times 15$ for subpixel matching with edge filtering based on $11 \times 11$ window in images of size $1024 \times 768$ and a disparity range of $115$ pixels. We filter the matching result with {\em back-and-forth} verification and a size constraint on the matches in a neighborhood of $200$ pixels to give us an incomplete point cloud with mainly reliable matches. We segment the point cloud based on the color image of the reference camera. We adapted the code of Horprasert et al.~\cite{Horprasert99} for background subtraction in our controlled lighting environment, for which we capture a series of background images without the deformable object to be captured. After extracting the foreground object, we estimate normals based on the point cloud in the image domain. We slide a window of $5 \times 5$ pixels over the image and fit a least-squares plane to the $\{x,y,z\}$-coordinates using singular value decomposition. If the singular values indicate that the plane approximation is poor or if less than $50\%$ plus $1$ depth values are available in the current window, we discard the current stereo match as an outlier. This gives us a filtered 3D point cloud with normal vectors at every point.

\begin{figure*}[tbp]
\centering
\begin{tabular}{|cc|ccc|}
\hline
\includegraphics[height = 1.5cm]{ear_cropped_data_colored20.jpg} &
\includegraphics[height = 1.3cm]{ear_template.jpg} & 
\includegraphics[height = 2.3cm]{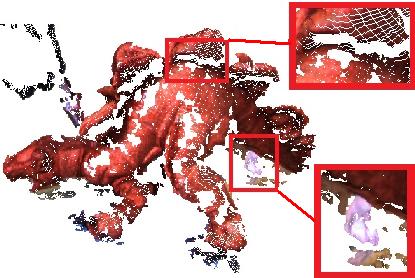} &
\includegraphics[height = 1.8cm]{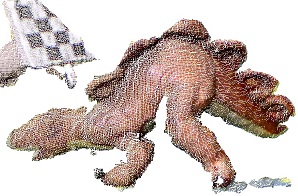} &
\includegraphics[height = 1.6cm]{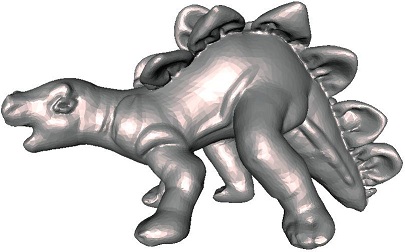} \\
\hline
\end{tabular}
\caption{\textit{\small Left: part of noisy stereo frame and template for ear model. Right: part of noisy stereo frame, part of Kinect frame, and template for dinosaur model.}}
\label{ear_dino_template}
\end{figure*}

Our experiments use two datasets acquired using this setup. The first dataset is a silicon ear used for acupuncture training that is acquired while the helix of the ear is being pushed down. Fig.~\ref{ear_dino_template} shows one of the input frames and the template mesh (containing 19993 vertices). The second dataset is a dinosaur plush toy that is acquired in two sequences: first while a flap of the model is being pushed towards the spine and second while the neck of the model is being pushed towards the floor. Fig.~\ref{ear_dino_template} shows one of the input frames and the template mesh (containing 25315 vertices). Note that all of these datasets contain noise that is typical for data acquired using stereo algorithms, such as noise along the viewing direction and missing data due to occlusions. Occlusions are especially visible in the areas of the helix of the ear model and the flaps of the dinosaur model. Furthermore, the input data of the dinosaur model contains points located on a tag attached to the tail of the model. This tag is not part of the template model, and handling this discrepancy is challenging. 

\textbf{Range Data.} We acquire range data using a Kinect sensor and the point cloud library. With this setup, we acquire a deformation of the dinosaur plush toy that is similar to one of the sequences acquired using the stereo camera. Furthermore, we acquire a deformation where the tail of the dinosaur toy is pushed to the side. The resolution of this type of data is low compared to our stereo data, as shown in Fig.~\ref{ear_dino_template}.

\textbf{Force Probe.} The position and orientation of the force sensor probe can be tracked by fixing a checkerboard pattern to the probe handle, and by tracking the corners of the checkerboard. The resulting position and orientation of the force sensor can be used to remove the data caused by the sensor from the point cloud. We take advantage of this option for the ear model, where the probe has a similar scale as the small-scale geometry of the helix. For all remaining datasets, the information on the location of the force sensor is not used.

\textbf{Template Model.} For all synthetic datasets, we use the complete undeformed model as template. For the ear and dinosaur models, we generate a template by acquiring the geometry of the model in a rest pose using a 3D scanner. The different views of the scans are merged and processed manually to lead to a watertight mesh. 

\subsection{Evaluation of Robustness with respect to Noise}

\begin{figure*}[tbp]
\centering
\begin{tabular}{|c||c|c||c|c||c|c||c|c|}
\hline
{\small Frame} & \multicolumn{2}{|c||}{\small No Noise} & \multicolumn{2}{|c||}{\small Outliers} & \multicolumn{2}{|c||}{\small Gaussian Noise} & \multicolumn{2}{|c|}{\small Resolution} \\
\hline
& {\small Input} & {\small Output} & {\small Input} & {\small Output} & {\small Input} & {\small Output} & {\small Input} & {\small Output} \\
\hline
\multirow{1}{*}[0.7cm]{\small 1} & \includegraphics[scale=0.25]{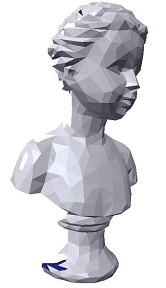} & \includegraphics[scale=0.25]{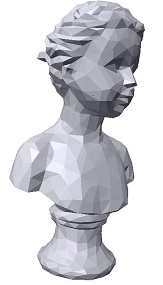} & \includegraphics[scale=0.25]{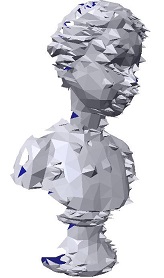} & \includegraphics[scale=0.25]{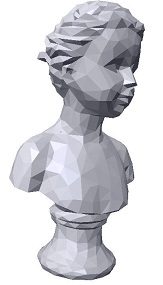} & \includegraphics[scale=0.25]{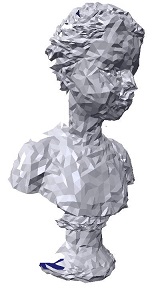} & \includegraphics[scale=0.25]{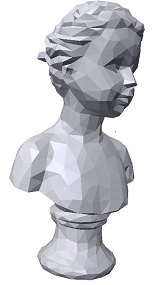} & \includegraphics[scale=0.25]{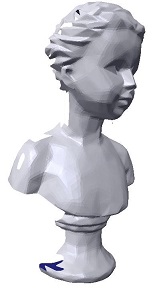} & \includegraphics[scale=0.25]{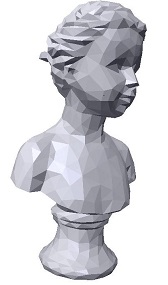}\\
\hline
\multirow{1}{*}[0.7cm]{\small 4} & \includegraphics[scale=0.25]{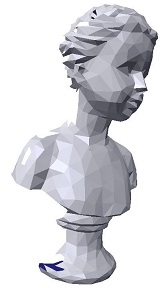} & \includegraphics[scale=0.25]{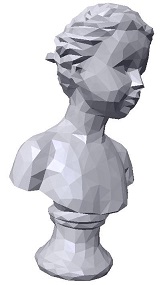} & \includegraphics[scale=0.25]{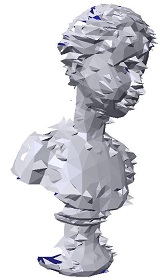} & \includegraphics[scale=0.25]{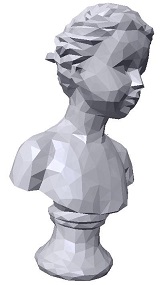} & \includegraphics[scale=0.25]{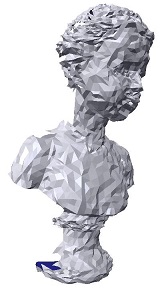} & \includegraphics[scale=0.25]{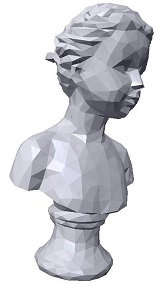} & \includegraphics[scale=0.25]{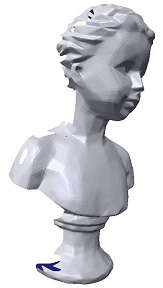} & \includegraphics[scale=0.25]{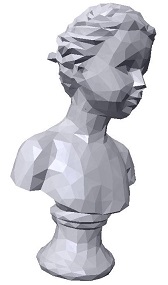}\\
\hline
\multirow{1}{*}[0.7cm]{\small 7} & \includegraphics[scale=0.25]{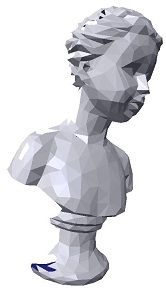} & \includegraphics[scale=0.25]{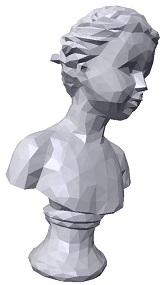} & \includegraphics[scale=0.25]{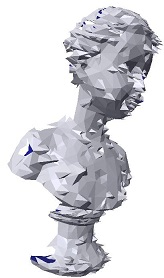} & \includegraphics[scale=0.25]{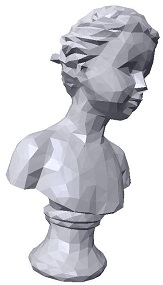} & \includegraphics[scale=0.25]{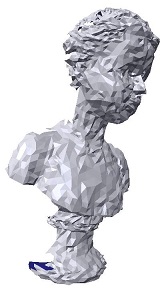} & \includegraphics[scale=0.25]{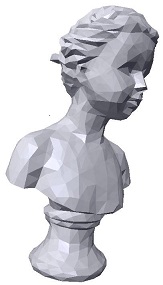} & \includegraphics[scale=0.25]{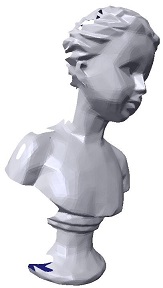} & \includegraphics[scale=0.25]{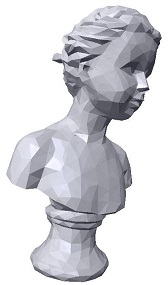}\\
\hline
\multirow{1}{*}[0.7cm]{\small 10} & \includegraphics[scale=0.25]{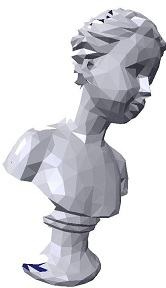} & \includegraphics[scale=0.25]{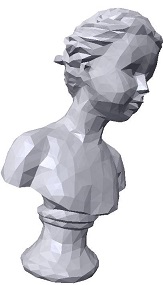} & \includegraphics[scale=0.25]{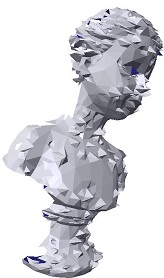} & \includegraphics[scale=0.25]{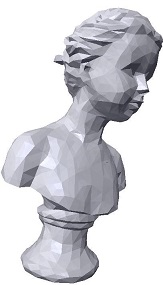} & \includegraphics[scale=0.25]{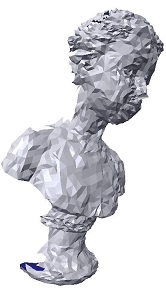} & \includegraphics[scale=0.25]{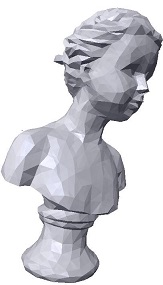} & \includegraphics[scale=0.25]{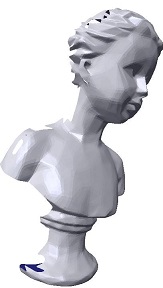} & \includegraphics[scale=0.25]{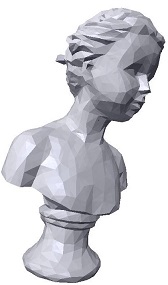}\\
\hline
\end{tabular}
\caption{\textit{\small Synthetic noise evaluation. Each column shows the input data and the results of our method.}}
\label{buste_error_visualization}
\end{figure*}

\begin{figure*}[tbp]
\centering
\begin{tabular}{lll}
Mean Distance & Maximum Distance & \\
\includegraphics[height = 4.0cm]{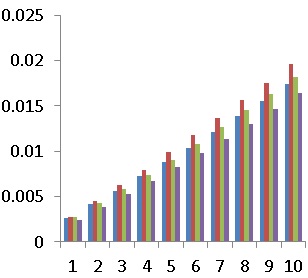} & 
\includegraphics[height = 4.0cm]{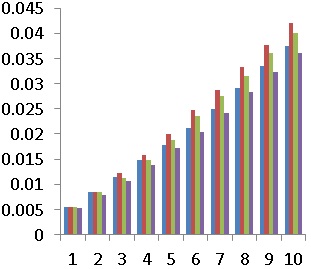} &
\includegraphics[height = 4.0cm]{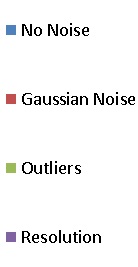}\\
\end{tabular}
\caption{\textit{\small Errors of the deformations shown in Fig.~\ref{buste_error_visualization} measured as mean and maximum distance over all vertices. X-axis: frame number. Y-axis: distances (for reference: normalized mesh diagonal of length one).}}
\label{buste_error_measure}
\end{figure*}

Our first set of experiments aims to show that our approach is robust with respect to noise. To start, consider the synthetic bust data generated with a linear FEM. For this data, the complete ground truth model is known, and can therefore be used to quantitatively evaluate the robustness of our approach with respect to noise. In this experiment, we consider three types of noise. First, we aim to model outliers in a way that simulates the outliers commonly present in stereo data. To model these outliers, we pick a viewpoint for the model. A vertex $p_i$ is perturbed as $p_i + x \vec{v}_i$, where $\vec{v}_i$ is the unit vector pointing from $p_i$ to the viewpoint and $x$ is a uniformly distributed random number in the range $\left[-r,4r\right]$, and $r$ is the resolution of the model. Each vertex of the model is perturbed with probability 1/10. Second, we perturb the vertices of the input data by adding Gaussian noise in the vertices normal direction. The variance of the Gaussian is $2\%$ of the bounding ball radius of the model. Third, we evaluate the influence of the resolution of the input data on the results by subdividing the input data using one step of Loop subdivision.

\begin{figure*}[tbp]
\centering
\begin{tabular}{|cccccc|}
\hline
\multicolumn{6}{|l|}{Ear Sequence (Stereo Data)}\\
\includegraphics[width = 0.1\textwidth]{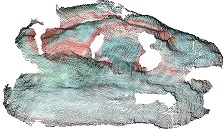} & 
\includegraphics[width = 0.1\textwidth]{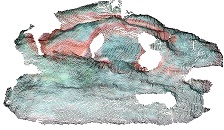} &
\includegraphics[width = 0.1\textwidth]{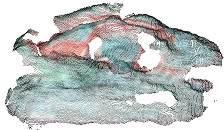} &
\includegraphics[width = 0.1\textwidth]{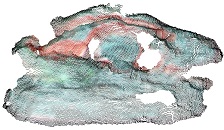} &
\includegraphics[width = 0.1\textwidth]{ear_cropped_data_colored20.jpg} & \\
\includegraphics[width = 0.1\textwidth]{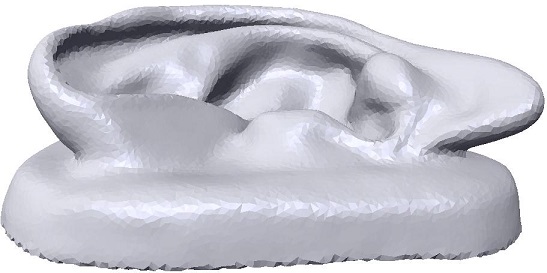} & 
\includegraphics[width = 0.1\textwidth]{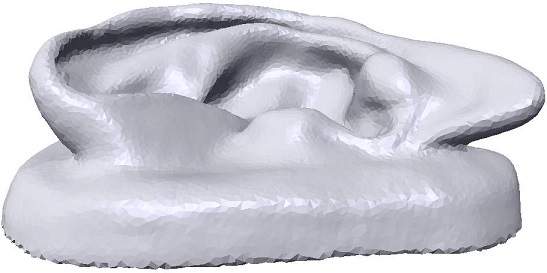} &
\includegraphics[width = 0.1\textwidth]{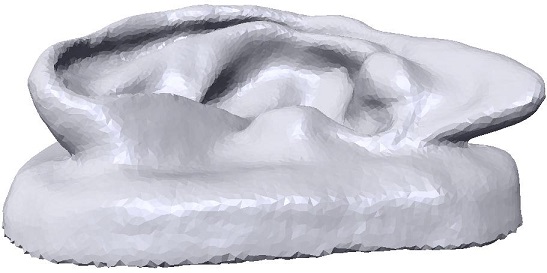} &
\includegraphics[width = 0.1\textwidth]{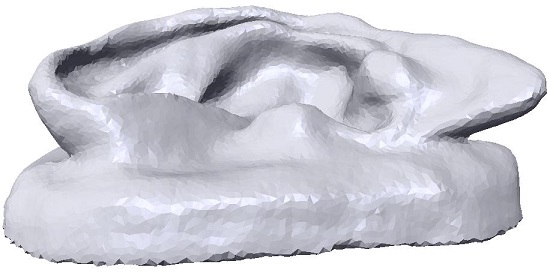} &
\includegraphics[width = 0.1\textwidth]{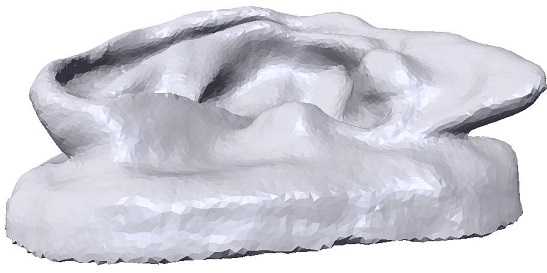} & 
\multirow{2}{*}[1.5cm]{\includegraphics[width = 0.045\textwidth]{ear_color_legend.jpg}}\\
\includegraphics[width = 0.1\textwidth]{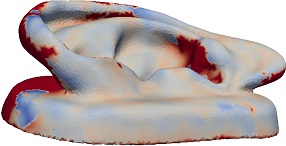} & 
\includegraphics[width = 0.1\textwidth]{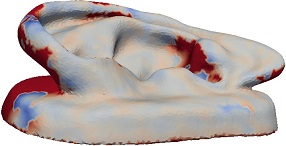} &
\includegraphics[width = 0.1\textwidth]{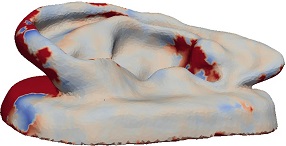} &
\includegraphics[width = 0.1\textwidth]{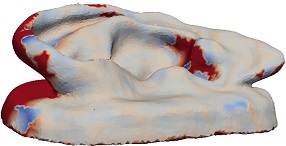} &
\includegraphics[width = 0.1\textwidth]{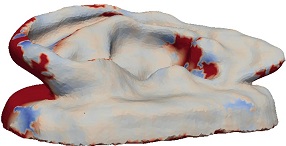} & \\
\hline
\multicolumn{6}{|l|}{Dinosaur Flap Sequence (Stereo Data)}\\
\includegraphics[width = 0.1\textwidth]{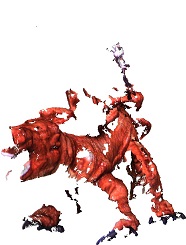} & 
\includegraphics[width = 0.1\textwidth]{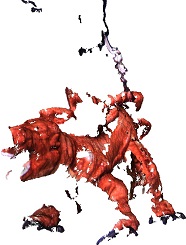} &
\includegraphics[width = 0.1\textwidth]{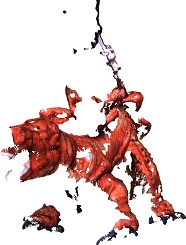} &
\includegraphics[width = 0.1\textwidth]{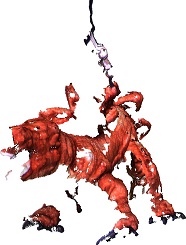} &
\includegraphics[width = 0.1\textwidth]{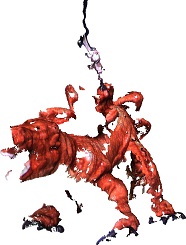} & \\
\includegraphics[width = 0.08\textwidth]{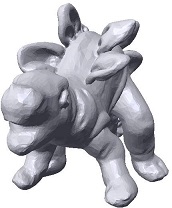} & 
\includegraphics[width = 0.08\textwidth]{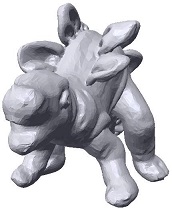} &
\includegraphics[width = 0.08\textwidth]{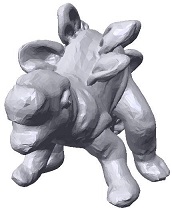} &
\includegraphics[width = 0.08\textwidth]{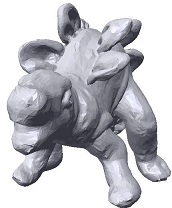} &
\includegraphics[width = 0.08\textwidth]{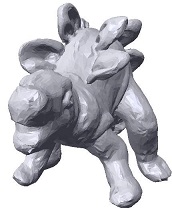} & 
\multirow{2}{*}[1.5cm]{\includegraphics[width = 0.045\textwidth]{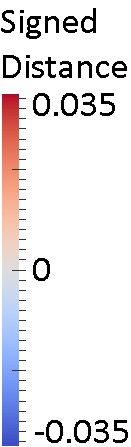}}\\
\includegraphics[width = 0.08\textwidth]{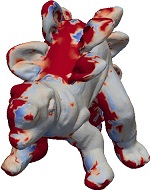} & 
\includegraphics[width = 0.08\textwidth]{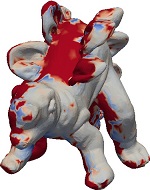} &
\includegraphics[width = 0.08\textwidth]{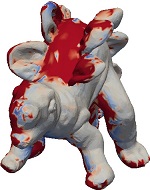} &
\includegraphics[width = 0.08\textwidth]{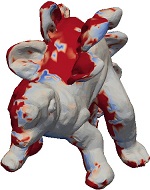} &
\includegraphics[width = 0.08\textwidth]{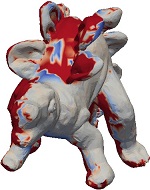} & \\
\hline
\multicolumn{6}{|l|}{Dinosaur Neck Sequence (Stereo Data)}\\
\includegraphics[width = 0.13\textwidth]{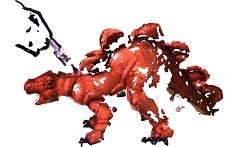} & 
\includegraphics[width = 0.13\textwidth]{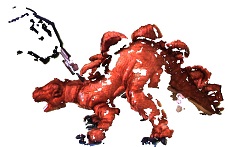} &
\includegraphics[width = 0.13\textwidth]{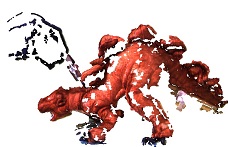} &
\includegraphics[width = 0.13\textwidth]{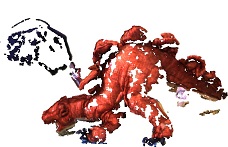} &
\includegraphics[width = 0.13\textwidth]{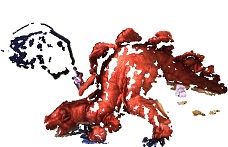} & \\
\includegraphics[width = 0.13\textwidth]{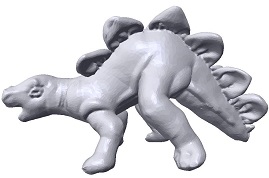} & 
\includegraphics[width = 0.13\textwidth]{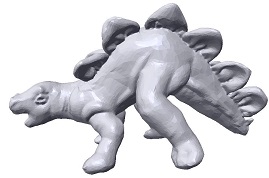} &
\includegraphics[width = 0.13\textwidth]{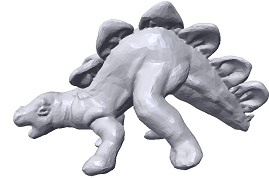} &
\includegraphics[width = 0.13\textwidth]{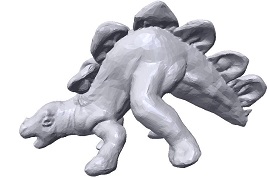} &
\includegraphics[width = 0.13\textwidth]{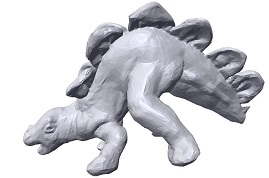} & 
\multirow{2}{*}[1.5cm]{\includegraphics[width = 0.045\textwidth]{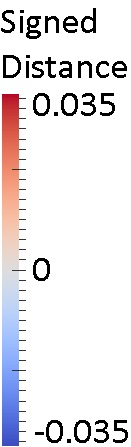}}\\
\includegraphics[width = 0.13\textwidth]{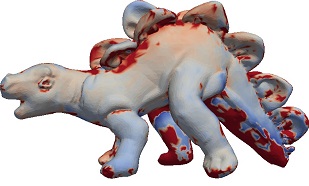} & 
\includegraphics[width = 0.13\textwidth]{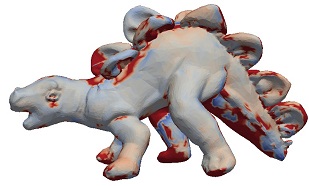} &
\includegraphics[width = 0.13\textwidth]{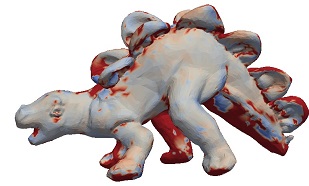} &
\includegraphics[width = 0.13\textwidth]{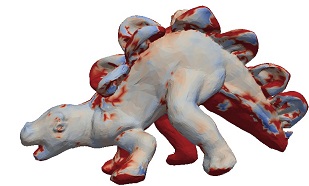} &
\includegraphics[width = 0.13\textwidth]{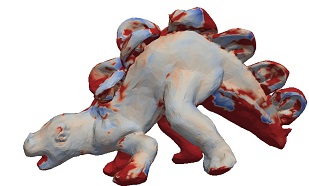} & \\
\hline
\multicolumn{6}{|l|}{Dinosaur Neck Sequence (Range Data)}\\
\includegraphics[width = 0.13\textwidth]{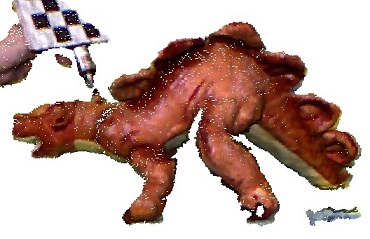} & 
\includegraphics[width = 0.13\textwidth]{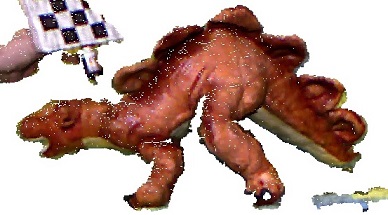} &
\includegraphics[width = 0.13\textwidth]{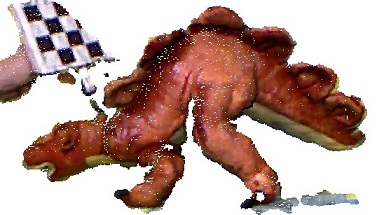} &
\includegraphics[width = 0.13\textwidth]{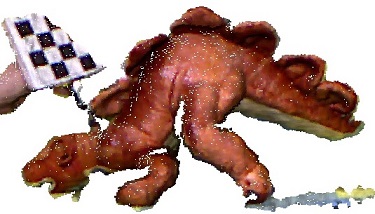} &
\includegraphics[width = 0.13\textwidth]{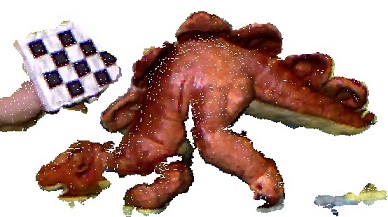} & \\
\includegraphics[width = 0.13\textwidth]{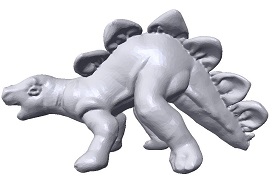} & 
\includegraphics[width = 0.13\textwidth]{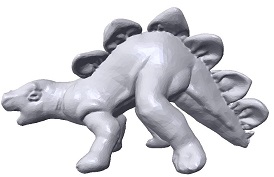} &
\includegraphics[width = 0.13\textwidth]{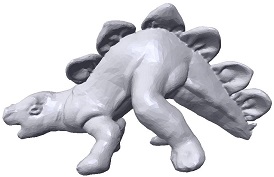} &
\includegraphics[width = 0.13\textwidth]{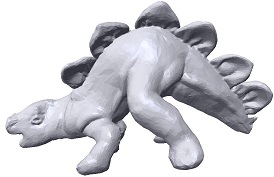} &
\includegraphics[width = 0.13\textwidth]{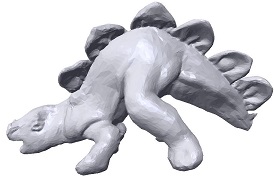} & 
\multirow{2}{*}[1.5cm]{\includegraphics[width = 0.045\textwidth]{neck_color_legend.jpg}}\\
\includegraphics[width = 0.13\textwidth]{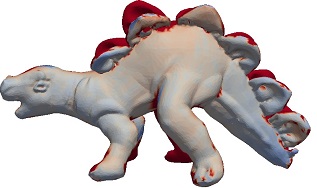} & 
\includegraphics[width = 0.13\textwidth]{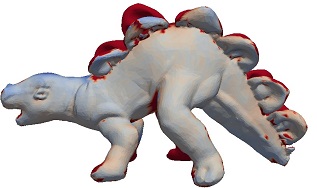} &
\includegraphics[width = 0.13\textwidth]{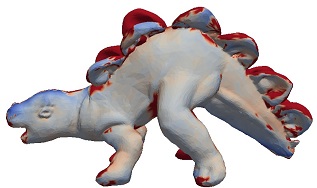} &
\includegraphics[width = 0.13\textwidth]{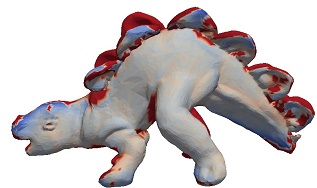} &
\includegraphics[width = 0.13\textwidth]{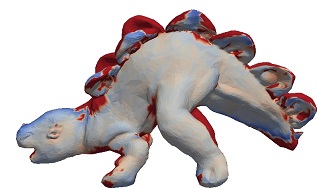} & \\
\hline
\end{tabular}
\caption{\textit{\small Observed sides of the acquired datasets. For each example, the figure shows from top to bottom the data, the result, and the signed distance between result and data (viewpoints may differ slightly).}}
\label{tracking_observed_sides}
\end{figure*}

Fig.~\ref{buste_error_visualization} shows that the results obtained by our method are not visually affected by noise. As the per-vertex distance for a frame can be computed as the distance between the vertex position in the tracking result and its corresponding point in the ground truth model, the error for a frame can then be computed as the average or maximum over all per-vertex distances of this frame. The resulting errors are given in Fig.~\ref{buste_error_measure}. The average distances are small compared to the size of the model and the increase in the distances caused by the presence of outliers, Gaussian noise, and an increase in resolution is insignificant. In our template-based tracking, distances increase linearly over time.

Second, we qualitatively evaluate the results of tracking noisy stereo and range datasets. Fig.~\ref{tracking_observed_sides} shows the observed sides of four sequences acquired using either a stereo or a range camera. The figure shows the data, the deformed template, and the signed distance between the deformed template and the data. In the visualization of the signed distance, points that do not have a valid nearest neighbor in the data are shown in red. Note that the tracking result captures the overall deformation of the models in spite of large noise, high levels of occlusion, and additional data. Furthermore, it is noteworthy that results of similar quality are obtained for stereo and range data, which have significantly different resolution and noise levels. An additional result for a deformation of the dinosaur model, where the tail is pushed to the side, is shown in the supplementary material.

\begin{figure}[tb]
\centering
\includegraphics[width = 0.48\textwidth]{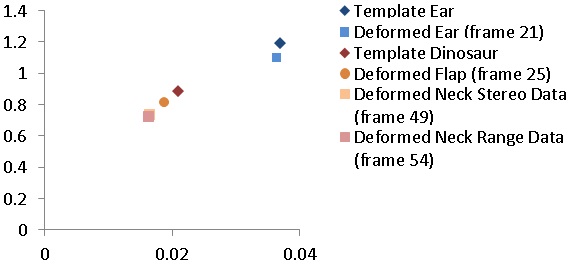}
\caption{\textit{\small Influence of the deformations shown in Fig.~\ref{tracking_observed_sides} on the geometric measures volume ($x$-axis) and surface area ($y$-axis). For reference: normalized
mesh diagonal of length one.}}
\label{tracking_real_surface_measures}
\end{figure}

To quantitatively evaluate how well the shape of the models is preserved during the deformation, we measure the volume and the surface area of the template models and of the deformed template at the last frames of the deformation sequences. Fig.~\ref{tracking_real_surface_measures} shows that volume and surface area are preserved well throughout the deformation sequences for all four real-world datasets. Since all of these deformations are approximately volume- and area-preserving in reality, this result gives supporting evidence that the method performs well.

\subsection{Evaluation of FEM Correction}

Our second set of experiments shows that the FEM step improves the shape of the unseen side of the model. We show that displacing the unobserved vertices using a linear FEM in every deformation step yields satisfactory results even for synthetic deformations that were generated using a non-linear FEM and for real-world deformation sequences.

\subsubsection*{Comparison to Ground Truth Deformations}

\begin{figure*}[tb]
\centering
\begin{tabular}{l l c}
Bust & Hand & \\
\includegraphics[height=5.0cm]{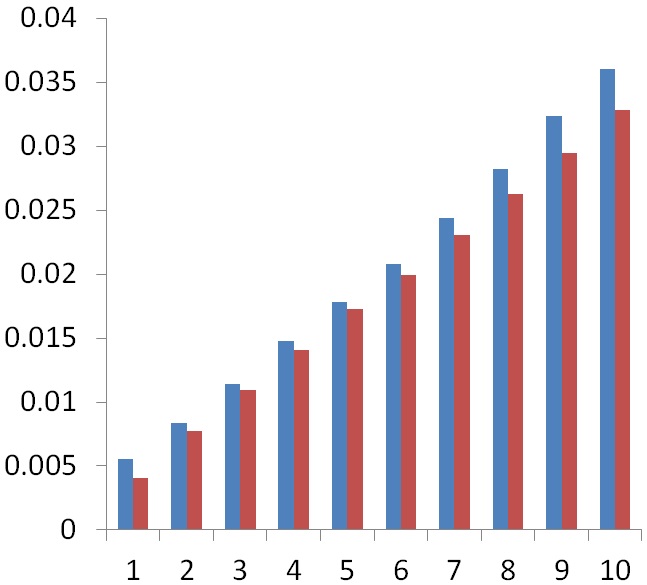} & 
\includegraphics[height=5.0cm]{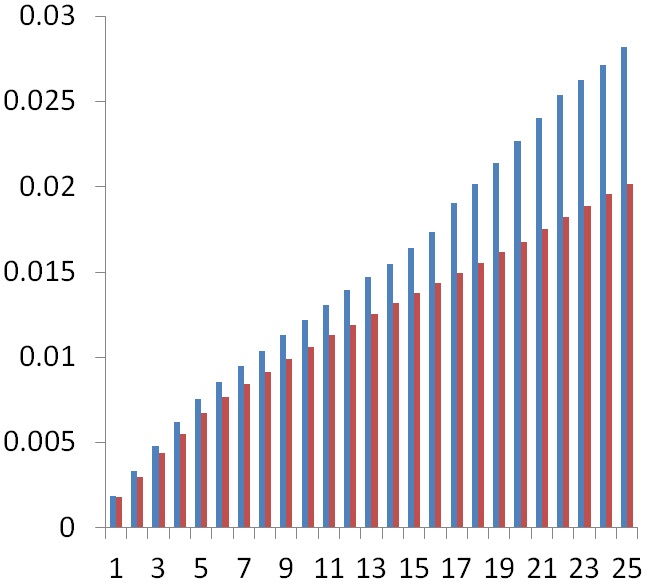} &
\multirow{1}{*}[3.5cm]{\includegraphics[height=0.9cm]{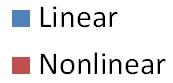}}\\
\end{tabular}
\caption{\textit{\small Errors measured as maximum distance over unobserved vertices. X-axis: frame number. Y-axis: distances (for reference: normalized mesh diagonal of length one).}}
\label{comp_fem_max_dist}
\end{figure*}

We consider the synthetic deformation sequences of the bust and hand models generated using linear FEM and the StVK model. For each deformation sequence, tracking is computed using our algorithm. We then evaluate the errors as above. However, instead of considering all vertices of the model for the error computation, we only consider vertices on the unseen side. The bust dataset exhibits a deformation that affects the global shape of the model, while the hand dataset exhibits the strongest deformation locally on one finger. Fig.~\ref{comp_fem_max_dist} shows that linear and non-linear deformations as well as deformations that affect the global shape and deformations that affect the local shape are tracked equally well.

For the models generated using linear FEM, we can also compare the estimated material parameters to the ground truth. In our experiments, the estimated Young's modulus is stable across the deformation sequence and can be obtained from the ground truth by a scaling. This is to be expected, as we normalize the lengths of the force directions. The estimated Poisson ratio varies across the input frames and on average, the true Poisson ratio is underestimated by approximately 0.25.

\subsubsection*{Comparison to Global Linear FEM}

\begin{figure}[tb]
\centering
\begin{tabular}{c}
Results for end position \\ 
\includegraphics[width=0.3\textwidth]{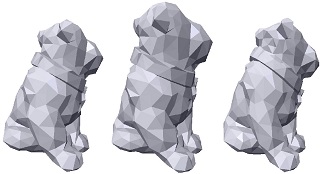} \\
Volume \\
\includegraphics[width=0.4\textwidth]{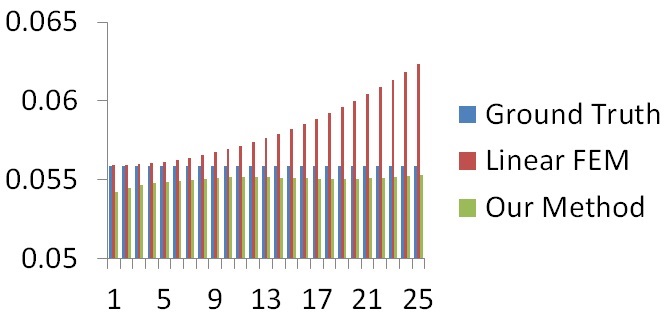} \\
\end{tabular}
\caption{\textit{\small Bulldog results. Top (left to right): ground truth, result computed using a global linear FEM, our result. Bottom: ground truth frames and volume of model (with normalized size) measured per frame.}}
\label{duck_eval}
\end{figure}

We compare the result of our method to a deformation computed by a global linear FEM. This experiment uses the synthetic bulldog data and is shown in Fig.~\ref{duck_eval}. The figure shows the result of applying a linear FEM deformation to the bulldog model and our result. The global linear FEM aims to linearize the global deformation that is observed when applying a non-linear FEM to the model. Observe that applying a linear FEM leads to unrealistic artifacts at the right front leg and the head of the bulldog because the applied force causes a rotation. Recall that our method uses a linear FEM to predict the unobserved side of the model. However, since our method linearizes the deformation locally at each frame, no unrealistic artifacts occur. Furthermore, our approach is significantly closer to the ground truth than using a global linear FEM in terms of mesh volume and area.

\subsection{Comparison to Surface-Based Deformation}

We compare the results of our method to the results of only using the surface-based template deformation method outlined in Section~\ref{sec:tracking}. Choosing this surface-based deformation technique results in deformations that predict the unobserved side using a term that aims to preserve a smooth deformation field. Comparing to this technique directly evaluates the influence of the FEM step on the result. In the following, we refer to the surface-based template deformation as our method without FEM.

\begin{figure}[tb]
\centering
\includegraphics[height=5.0cm]{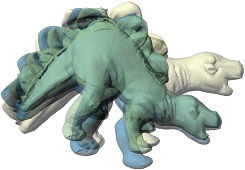}
\caption{\textit{\small Neck results for the unobserved side. The template is shown in yellow, the surface-based deformation result in blue, and our result in green.}}
\label{neck_fem_eval}
\end{figure}

We evaluate the influence of the FEM correction for tracking noisy scans. Consider the neck sequence acquired using the stereo setup. Fig.~\ref{neck_fem_eval} shows the template (yellow) and the results at the end of the sequence with (green) and without (blue) the FEM correction. Note how the surface-based deformation finds a solution that deforms the template smoothly, which leads to a translation of the leg rather than a bending. Furthermore, the tail is merely translated upwards. Using our method, the legs slide and bend realistically, and the model’s tail lifts up, as in reality.

\begin{figure*}[t]
\centering
\begin{tabular}{|c|c|c|c| l l}
\cline{1-4}
Frame & Input & Wuhrer et al. & Ours & Mean Distance & Max. Distance\\
& & (2012) & &
\includegraphics[height = 0.6cm]{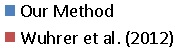} & \\
\cline{1-4}
\multirow{1}{*}[0.7cm]{1} & \includegraphics[scale=0.25]{buste_linear_data_0.jpg} & \includegraphics[scale=0.25]{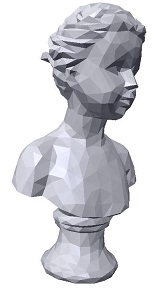} & \includegraphics[scale=0.25]{buste_linear_0.jpg} & 
\multirow{4}{*}[1.5cm]{\includegraphics[height = 4.0cm]{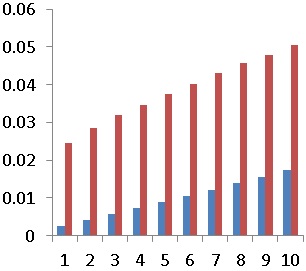}} & 
\multirow{4}{*}[1.5cm]{\includegraphics[height = 4.0cm]{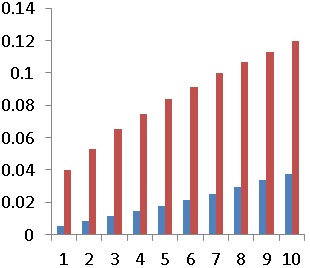}} \\
\cline{1-4}
\multirow{1}{*}[0.7cm]{4} & \includegraphics[scale=0.25]{buste_linear_data_3.jpg} & \includegraphics[scale=0.25]{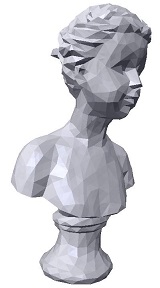} & \includegraphics[scale=0.25]{buste_linear_3.jpg} & & \\
\cline{1-4}
\multirow{1}{*}[0.7cm]{7} & \includegraphics[scale=0.25]{buste_linear_data_6.jpg} & \includegraphics[scale=0.25]{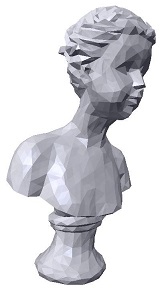} & \includegraphics[scale=0.25]{buste_linear_6.jpg} & & \\
\cline{1-4}
\multirow{1}{*}[0.7cm]{10} & \includegraphics[scale=0.25]{buste_linear_data_9.jpg} & \includegraphics[scale=0.25]{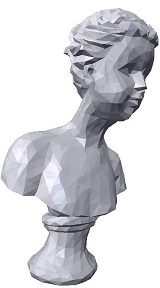} & \includegraphics[scale=0.25]{buste_linear_9.jpg} & & \\
\cline{1-4}
\end{tabular}
\caption{\textit{\small Synthetic comparison to Wuhrer et al. (2012). Left: input data and results. Right: distances to ground truth.}}
\label{buste_comp_3dimpvt}
\end{figure*}

Fig.~\ref{ear_comparison} (third from left) shows that the FEM correction also leads to a more physically plausible result for the ear model acquired using the stereo setup. In this case, using the FEM correction prevents a fattening of the object, which is mostly visible at the base and the helix.

\subsection{Comparison to Wuhrer et al. (2012)}

Finally, we compare the proposed method to its predecessor~\cite{wuhrer_lang_shu_3dimpvt12} and demonstrate that the proposed changes yield a significant improvement in the performance of the algorithm. To start, consider the synthetic bust model generated with linear FEM. We track this sequence using both the method by Wuhrer et al. (2012) and our method, and measure the mean and maximum distances over all vertices to the ground truth. Fig.~\ref{buste_comp_3dimpvt} shows the tracking results and the measured distances. Note that the mean and maximum distances of our method are less than a third of the corresponding distances of the method by Wuhrer et al. (2012). As can be seen in the left of the figure, this improvement is obtained because our method tracks the rotation of the head better than the method by Wuhrer et al. (2012) and because the back side of the bust is not flattened by our method.

\begin{figure*}[tp]
\centering
\begin{tabular}{|cccc|}
\hline
\multicolumn{4}{|l|}{Ear Sequence (Stereo Data)}\\
Input & Wuhrer et al. (2012) & Our Method & \\
\includegraphics[width=0.15\textwidth]{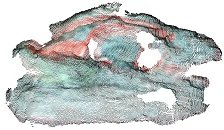} & 
\includegraphics[width=0.15\textwidth]{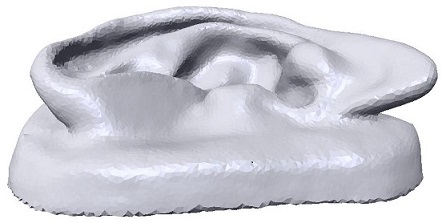} &
\includegraphics[width=0.15\textwidth]{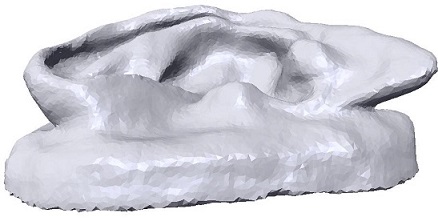} & \\
&
\includegraphics[width=0.15\textwidth]{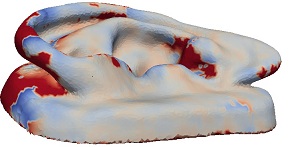} &
\includegraphics[width=0.15\textwidth]{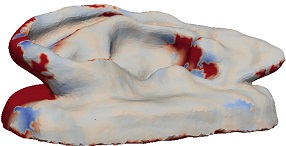} &
\multirow{1}{*}[2.0cm]{\includegraphics[height=2.2cm]{ear_color_legend.jpg}} \\
\hline
\multicolumn{4}{|l|}{Dinosaur Flap Sequence (Stereo Data)}\\
Input & Wuhrer et al. (2012) & Our Method & \\
\includegraphics[width=0.1\textwidth]{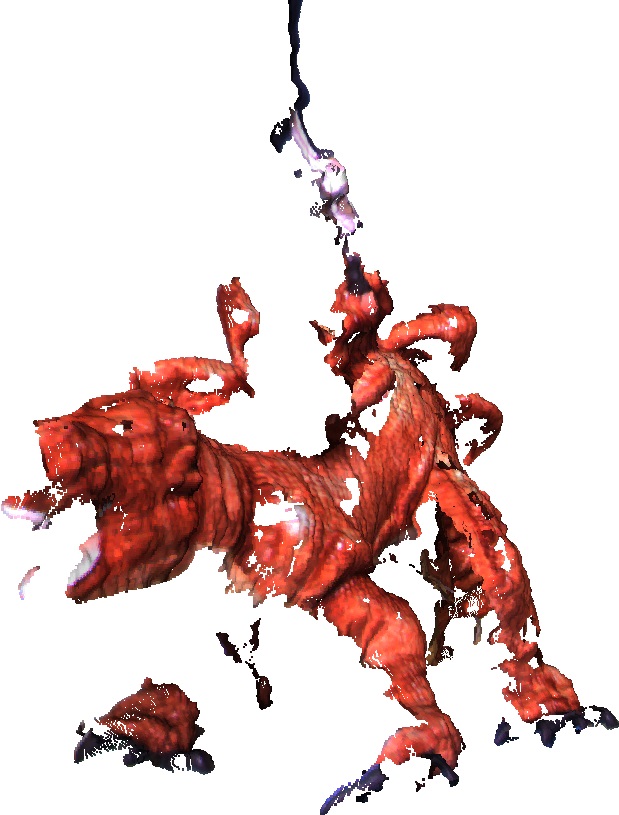} & 
\includegraphics[width=0.1\textwidth]{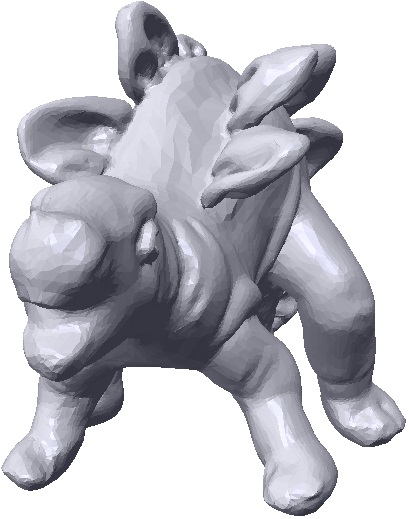} &
\includegraphics[width=0.1\textwidth]{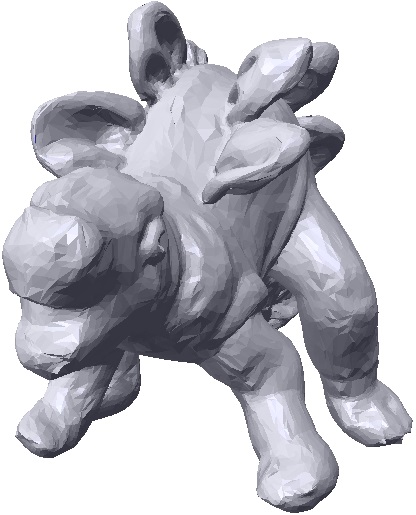} & \\
&
\includegraphics[width=0.1\textwidth]{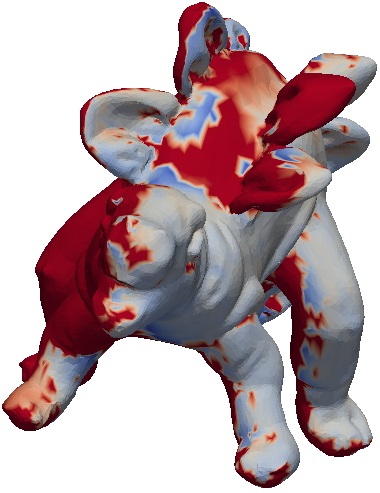} &
\includegraphics[width=0.1\textwidth]{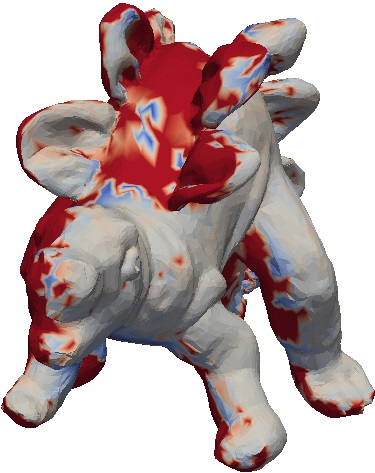} &  
\multirow{1}{*}[2.0cm]{\includegraphics[height=2.2cm]{flap_color_legend.jpg}} \\
\hline
\multicolumn{4}{|l|}{Dinosaur Neck Sequence (Stereo Data)}\\
Input & Wuhrer et al. (2012) & Our Method & \\
\includegraphics[width=0.2\textwidth]{neck_stereo_data_30.jpg} & 
\includegraphics[width=0.25\textwidth]{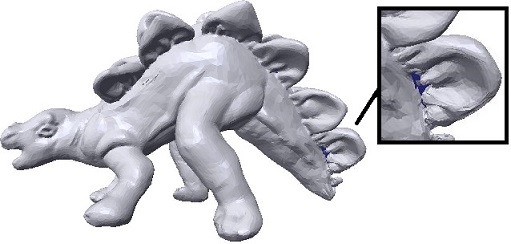} &
\includegraphics[width=0.25\textwidth]{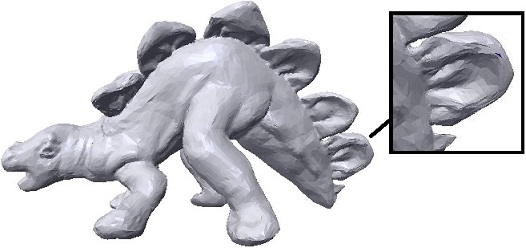} & \\
&
\includegraphics[width=0.2\textwidth]{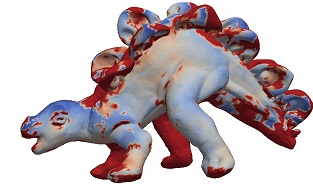} &
\includegraphics[width=0.2\textwidth]{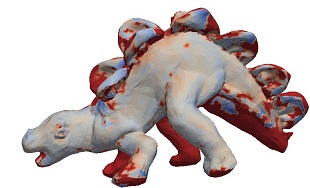} &  
\multirow{1}{*}[2.0cm]{\includegraphics[height=2.2cm]{neck_color_legend.jpg}} \\
\hline
\multicolumn{4}{|l|}{Dinosaur Neck Sequence (Range Data)}\\
Input & Wuhrer et al. (2012) & Our Method & \\
\includegraphics[width=0.2\textwidth]{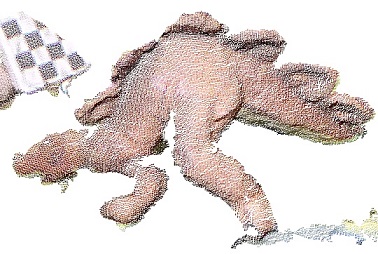} & 
\includegraphics[width=0.25\textwidth]{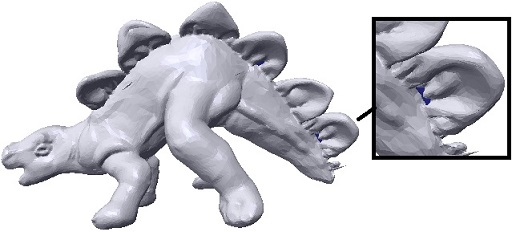} &
\includegraphics[width=0.25\textwidth]{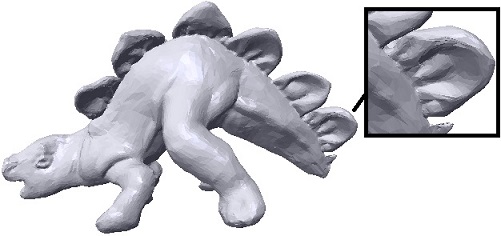} & \\
&
\includegraphics[width=0.2\textwidth]{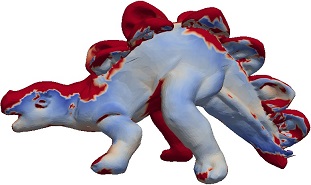} &
\includegraphics[width=0.2\textwidth]{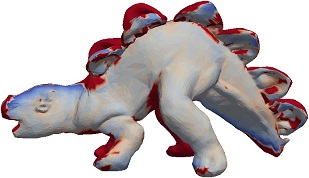} & 
\multirow{1}{*}[2.0cm]{\includegraphics[height=2.2cm]{neck_color_legend.jpg}} \\
\hline
\end{tabular}
\caption{\textit{\small Input data and results using the method by Wuhrer et al. (2012) and our method (also color-coded according to signed distance from data).}}
\label{real_data_comp_3dimpvt}
\end{figure*}

The improvement of our method over the method by Wuhrer et al. (2012) is especially noticeable for tracking noisy scan data. Fig.~\ref{ear_comparison} shows a comparison of the result obtained using the method by Wuhrer et al. (2012) (blue) to our result (green) for the last frame of the ear dataset acquired using a stereo setup. Note that while the method Wuhrer et al. (2012) does not track the local deformation of the helix, our method tracks the data correctly without resulting in a noisy output. Fig.~\ref{real_data_comp_3dimpvt} shows the results of the two methods for different frames of datasets acquired using stereo or range cameras. Note that our method results in less noise and self-intersections of the model (especially visible in the areas of the flaps of the dinosaur models), while at the same time yielding a higher data accuracy. This significant improvement is a consequence of the improvements of both the tracking step and the FEM prediction.

\subsection{Limitations}

The proposed method is currently designed to deform models of homogeneous isotropic material that are deformed by applying external forces on a small number of points. Modeling complex force fields or force fields acting on heterogeneous material, such as deformations of tissue caused by muscle movements, would require a segmentation of the model into regions of homogeneous materials and the input of a full force field. This is too tedious to acquire to be of practical use. However, by combining template-based tracking with simple physical simulation models, we make a first step in the direction of acquiring the geometry and material properties of an object jointly. 

We have demonstrated that our method, which uses a linear FEM model, allows to track both linear and non-linear material deformations accurately. This flexibility comes at the cost of material parameter estimates that vary over time and are not expected to be physically meaningful. In the future, we plan to explore modeling non-linear material behaviour explicitly, and to find stable and physically meaningful material parameters for this scenario by considering all available frames for material parameter estimation.

\begin{figure}[tb]
\centering
\includegraphics[height=3.0cm]{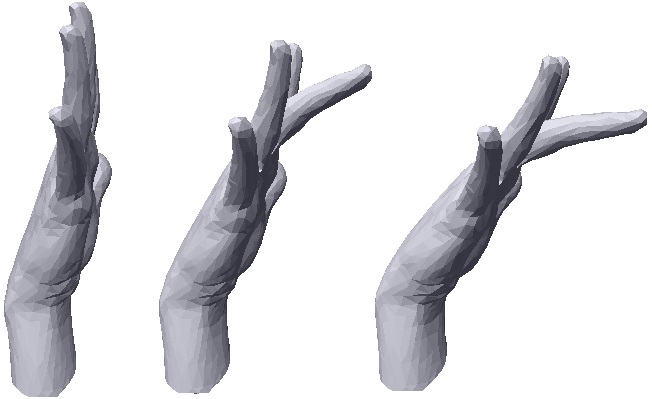}
\includegraphics[height=3.0cm]{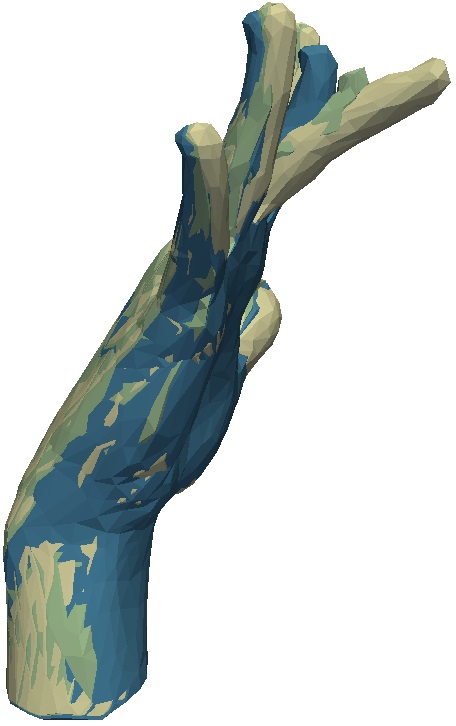}
\caption{\textit{\small Left: start frame, intermediate frame, and end frame of the input data. Right: result at the intermediate frame (shown in yellow) when the sparsely sampled simulation is used (shown in blue) and when the densely sampled simulation is used (shown in green).}}
\label{hand_large_def}
\end{figure}

As our approach employs a non-rigid iterative closest point algorithm to fit the template to the data, tracking large deformations may lead to drift. An example of this problem is included in the supplementary material. Furthermore, due to the non-rigid iterative closest point algorithm, our method cannot deform the template accurately if the initial alignment is poor or if there is significant deformation between consecutive frames. Hence, in cases of extreme deformations that are sampled sparsely in time, our tracking may get lost. This is shown in Fig.~\ref{hand_large_def}. Here, we simulated the same hand deformation twice; once sampled sparsely in time using 50 frames, and once sampled densely in time using 350 frames. For a particular frame (frame 20 in the sparsely sampled simulation, which corresponds to frame 140 in the densely sampled simulation), the ground truth deformation is shown in yellow, the result for tracking using the sparse sequence is shown in blue, and the result for tracking using the dense sequence is shown in green. Note that by using more frames, the tracking is able to follow the data more closely at the cost of additional drift.

\section{Conclusions}
\label{sec:conclusion}

We proposed an approach to track the geometry of a surface over time from a set of input point clouds captured from a single viewpoint. We combine the use of a template and the use of a linear finite element method to track the model. By linearizing the deformation at each frame, we show that we can accurately track surfaces that deform in a non-linear fashion. We demonstrate the robustness of our approach with respect to noise using a synthetic evaluation and using real data captured with a stereo setup and with a depth camera.

We leave the following ideas for future work. The tracking is lost when the distance between consecutive frames is large. This could potentially be addressed by tracking feature points on the model and by using these features to guide the non-rigid motion of the template during tracking. Furthermore, our approach assumes that a template is known a priori.  While this assumption is commonly used in 3D tracking approaches, it will be interesting to relax this requirement in the future. One way to relax this requirement would be to assume that the undeformed object is observed from a single moving viewpoint before the deformation, which allows to fuse these views into a template shape automatically.

\section*{Acknowledgements}

This work has partially been funded by NSERC, Canada, Networks of Centres of Excellence GRAND, Canada, and the Cluster of Excellence on \textit{Multimodal Computing and Interaction} within the Excellence Initiative of the German Federal Government.


{\small
\begin{thebibliography}{10}

\bibitem{aimAtShape}
Aim@Shape.
\newblock http://shapes.aimatshape.net/releases.php.

\bibitem{allen_curless_popovic_03_parametrization_body_shape}
Brett Allen, Brian Curless, and Zoran Popovi\'{c}.
\newblock The space of human body shapes: reconstruction and parameterization
  from range scans.
\newblock {\em ACM Transactions on Graphics}, 22(3):587--594, 2003.
\newblock Proceedings of SIGGRAPH.

\bibitem{becker_teschner_07}
Markus Becker and Matthias Teschner.
\newblock Robust and efficient estimation of elasticity parameters using the
  linear finite element method.
\newblock In {\em SimVis}, pages 15--28, 2007.

\bibitem{bickel_etal_10}
Bernd Bickel, Moritz B{\"{a}}cher, Miguel Otaduy, Hyunho~Richard Lee, Hanspeter
  Pfister, Markus Gross, and Wojciech Matusik.
\newblock Design and fabrication of materials with desired deformation
  behavior.
\newblock {\em ACM Transactions on Graphics}, 29(3), 2010.
\newblock Proceedings of SIGGRAPH.

\bibitem{bickel_etal_09}
Bernd Bickel, Moritz B{\"{a}}cher, Miguel Otaduy, Wojciech Matusik, Hanspeter
  Pfister, and Markus Gross.
\newblock Capture and modeling of non-linear heterogeneous soft tissue.
\newblock {\em ACM Transactions on Graphics}, 28(3), 2009.
\newblock Proceedings of SIGGRAPH.

\bibitem{bro-nielsen_fem_98}
Morten Bro-Nielsen.
\newblock Finite element modeling in surgery simulation.
\newblock {\em Proceedings of the IEEE}, 86:490--503, 1998.

\bibitem{cagniart2010CVPR}
Cedric Cagniart, Edmond Boyer, and Slobodan Ilic.
\newblock Free-from mesh tracking: a patch-based approach.
\newblock In {\em IEEE Conference on Computer Vision and Pattern Recognition},
  2010.

\bibitem{cagniart2010ECCV}
Cedric Cagniart, Edmond Boyer, and Slobodan Ilic.
\newblock Probabilistic deformable surface tracking from multiple videos.
\newblock In {\em European Conference on Computer Vision}, 2010.

\bibitem{choi_szymczak_09}
Jaeil Choi and Andrzej Szymczak.
\newblock Fitting solid meshes to animated surfaces using linear elasticity.
\newblock {\em ACM Transactions on Graphics}, 28(1):6:1--6:10, 2009.

\bibitem{deAguiar_et_al_08}
Edilson de~Aguiar, Carsten Stoll, Christian Theobalt, Naveed Ahmed, Hans-Peter
  Seidel, and Sebastian Thrun.
\newblock Performance capture from sparse multi-view video.
\newblock {\em ACM Transactions on Graphics}, 27(3):98:1--98:10, 2008.
\newblock Proceedings of SIGGRAPH.

\bibitem{dryden_mardia_shape_analysis}
Ian Dryden and Kanti Mardia.
\newblock {\em Statistical Shape Analysis}.
\newblock Wiley, 2002.

\bibitem{eskandari_etal_11_viscoelasticFEM}
Hani Eskandari, Septimiu Salcudean, Robert Rohling, and Ian Bell.
\newblock Real-time solution of the finite element inverse problem of
  viscoelasticity.
\newblock {\em Inverse Problems}, 27(8):085002:1--16, 2011.

\bibitem{furukawa_ponce_08}
Yasutaka Furukawa and Jean Ponce.
\newblock Dense 3d motion capture from synchronized video streams.
\newblock In {\em IEEE Conference on Computer Vision and Pattern Recognition},
  2008.

\bibitem{gao_etal_1996_elasticTissuesReview}
L.~Gao, K.~Parker, R.~Lerner, and S.~Levinson.
\newblock Imaging of the elastic properties of tissue--a review.
\newblock {\em Ultrasound in Medicine \& Biology}, 22:959--977, 1996.

\bibitem{garland_heckbert_97}
Michael Garland and Paul~S. Heckbert.
\newblock Surface simplification using quadric error metrics.
\newblock In {\em International Conference on Computer Graphics and Interactive
  Techniques}, pages 209--216, 1997.

\bibitem{getfem}
GetFEM.
\newblock http://download.gna.org/getfem/html/homepage/.

\bibitem{hensel_etal_2007_multiorganFEM}
Jennifer Hensel, Cynthia M\'{e}nard, Peter Chung, Michael Milosevic, Anna
  Kirilova, Joanne Moseley, Masoom Haider, and Kristy Brock.
\newblock Development of multiorgan finite element-based prostate deformation
  model enabling registration of endorectal coil magnetic resonance imaging for
  radiotherapy planning.
\newblock {\em International Journal of Radiation Oncology, Biology and
  Physics}, 68(5):1522--â€“1528, 2007.

\bibitem{Horprasert99}
Thanarat Horprasert, David Harwood, and Larry~S. Davis.
\newblock A statistical approach for real-time robust background subtraction
  and shadow detection.
\newblock In {\em ICCV Frame-Rate Workshop}, pages 1--19, 1999.

\bibitem{lang_pai_01}
Jochen Lang and Dinesh Pai.
\newblock Estimation of elastic constants from 3d range-flow.
\newblock In {\em 3D Digital Imaging and Modeling, International Conference
  on}, page 331, 2001.

\bibitem{lee-tmi_2012}
Huai-Ping Lee, Mark Foskey, Marc Niethammer, Pavel Krajcevski, and Ming Lin.
\newblock Simulation-based joint estimation of body deformation and elasticity
  parameters for medical image analysis.
\newblock {\em IEEE Transactions on Medical Imaging}, 31(11):2156--2168, 2012.

\bibitem{li09robust}
Hao Li, Bart Adams, Leonidas~J. Guibas, and Mark Pauly.
\newblock Robust single-view geometry and motion reconstruction.
\newblock {\em ACM Transactions on Graphics (Proceedings SIGGRAPH Asia 2009)},
  28(5), 2009.

\bibitem{li12corres}
Hao Li, Linjie Luo, Daniel Vlasic, Pieter Peers, Jovan Popovi\'{c}, Mark Pauly,
  and Szymon Rusinkiewicz.
\newblock Temporally coherent completion of dynamic shapes.
\newblock {\em ACM Transactions on Graphics}, 31(1):2:1--11, 2012.

\bibitem{li08corres}
Hao Li, Robert~W. Sumner, and Mark Pauly.
\newblock Global correspondence optimization for non-rigid registration of
  depth scans.
\newblock {\em Computer Graphics Forum}, 27(5):1421--1430, 2008.
\newblock Proceedings of SIGGRAPH.

\bibitem{liao_etal_2009}
Miao Liao, Qing Zhang, Huamin Wang, Ruigang Yang, and Minglun Gong.
\newblock Modeling deformable objects from a single depth camera.
\newblock In {\em IEEE International Conference on Computer Vision}, 2009.

\bibitem{liu_nocedal_lbfgsb}
Dong~C. Liu and Jorge Nocedal.
\newblock On the limited memory method for large scale optimization.
\newblock {\em Mathematical Programming}, 45:503--528, 1989.

\bibitem{mfoggp_dyn_reg_07}
Niloy~J. Mitra, Simon Flory, Maks Ovsjanikov, Natasha Gelfand, Leonidas Guibas,
  and Helmut Pottmann.
\newblock Dynamic geometry registration.
\newblock In {\em Symposium on Geometry Processing}, 2007.

\bibitem{nguyen_boyce_10}
Thao Nguyen and Brad Boyce.
\newblock An inverse finite element method for determining the anisotropic
  properties of the cornea.
\newblock {\em Biomechanics and Modeling in Mechanobiology}, To appear.

\bibitem{popa10corres}
Tiberiu Popa, Ian South-Dickinson, Derek Bradley, Alla Sheffer, and Wolfgang
  Heidrich.
\newblock Globally consistent space-time reconstruction.
\newblock {\em Computer Graphics Forum}, 29(5):1633--1642, 2010.
\newblock Proceedings of SGP.

\bibitem{rusinkiewicz_levoy_efficient_icp}
Szymon Rusinkiewicz and Marc Levoy.
\newblock Efficient variants of the icp algorithm.
\newblock In {\em Conference on 3D Digital Imaging and Modeling}, pages
  145--152, 2001.

\bibitem{schnabel_etal_03}
Julia Schnabel, Christine Tanner, Andy Castellano-Smith, Andreas Degenhard,
  Martin Leach, Rodney Hose, Derek Hill, and David Hawkes.
\newblock Validation of nonrigid image registration using finite-element
  methods: {A}pplication to breast {MR} images.
\newblock {\em IEEE Transactions on Medical Imaging}, 22(2):238--247, 2003.

\bibitem{sharf_et_al_08}
Andrei Sharf, Dan Alcantara, Thomas Lewiner, Chen Greif, Alla Sheffer, Nina
  Amenta, and Daniel Cohen-Or.
\newblock Space-time surface reconstruction using incompressible flow.
\newblock {\em ACM Transaction on Graphics}, 27(5), 2008.
\newblock Proceedings of Siggraph Asia.

\bibitem{syllebranque_boivin_08}
C\'{e}dric Syllebranque and Samuel Boivin.
\newblock Estimation of mechanical parameters of deformable solids from videos.
\newblock {\em The Visual Computer}, 24:963--972, 2008.

\bibitem{tam_survey_2013}
Gary Tam, Zhi-Quan Cheng, Yu-Kun Lai, Frank Langbein, Yonghuai Liu, David
  Marshall, Ralph Martin, Xian-Fang Sun, and Paul Rosin.
\newblock Registration of 3d point clouds and meshes: A survey from rigid to
  non-rigid.
\newblock {\em IEEE Transactions on Visualization and Computer Graphics},
  19:1199â€“--1217, 2013.

\bibitem{terzopoulos_82}
Demitri Terzopoulos.
\newblock Multi-level reconstruction of visual surfaces: Variational principles
  and finite element representations.
\newblock Technical Report AI Memo number 671, MIT, 1982.

\bibitem{tevs_etal_12}
Art Tevs, Alexander Berner, Michael Wand, Ivo Ihrke, Martin Bokeloh, Jens
  Kerber, and Hans-Peter Seidel.
\newblock Animation cartography - intrinsic reconstruction of shape and motion.
\newblock {\em ACM Transaction on Graphics}, 31(2):12:1--15, 2012.

\bibitem{tung_matsuyama_10}
Tony Tung and Takashi Matsuyama.
\newblock Dynamic surface matching by geodesic mapping for 3d animation
  transfer.
\newblock In {\em IEEE Conference on Computer Vision and Pattern Recognition},
  pages 1402--1409, 2010.

\bibitem{vanKaick_egstar10}
Oliver van Kaick, Hao Zhang, Ghassan Hamarneh, and Danial Cohen-Or.
\newblock A survey on shape correspondence.
\newblock In {\em Eurographics State-of-the-art Report}, 2010.

\bibitem{vlasic_etal_08}
Daniel Vlasic, Ilya Baran, Wojciech Matusik, and Jovan Popovic.
\newblock Articulated mesh animation from multi-view silhouettes.
\newblock {\em ACM Transactions on Graphics}, 27(3):97:1--10, 2008.
\newblock Proceedings of SIGGRAPH.

\bibitem{wand_et_al_07}
Michael Wand, Philipp Jenke, Qixing Huang, Martin Bokeloh, Leonidas Guibas, and
  Andreas Schilling.
\newblock Reconstruction of deforming geometry from time-varying point clouds.
\newblock In {\em Symposium on Geometry Processing}, 2007.

\bibitem{wuhrer_lang_shu_3dimpvt12}
Stefanie Wuhrer, Jochen Lang, and Chang Shu.
\newblock Tracking complete deformable objects with finite elements.
\newblock In {\em Conference on 3D Imaging Modeling Processing Visualization
  and Transmission}, 2012.

\bibitem{zheng_et_al_10}
Qian Zheng, Andrei Sharf, Andrea Tagliasacchi, Baoquan Chen, Hao Zhang, Alla
  Sheffer, and Daniel Cohen-Or.
\newblock Consensus skeleton for non-rigid space-time registration.
\newblock {\em Computer Graphcis Forum}, 29(2), 2010.
\newblock Proceedings of Eurographics.

\bibitem{zhu_etal_2003_youngsModulusReconstruction}
Yanning Zhu, Timothy~J. Hall, and Jingfeng Jiang.
\newblock A finite-element approach for youngâ€™s modulus reconstruction.
\newblock {\em IEEE Transactions on Medical Imaging}, 22(7):890--901, 2003.

\end{thebibliography}
}
\end{document}